\documentclass[a4paper,conference]{IEEEtran}

\usepackage{times}
\usepackage{epsfig}
\usepackage{graphicx}
\usepackage{amsmath}
\usepackage{amssymb}
\usepackage{siunitx}
\usepackage{subcaption}
\let\vec\mathbf
\begin{document}
%
\title{Can You Trust Your Pose? Confidence Estimation in Visual Localization}

\author{\IEEEauthorblockN{Luca Ferranti}
\IEEEauthorblockA{
University of Vaasa\\
Vaasa, Finland\\
luca.ferranti@univaasa.fi}
\and
\IEEEauthorblockN{Xiaotian Li}
\IEEEauthorblockA{Aalto University\\
Espoo, Finland\\
xiaotian.li@aalto.fi}
\and
\IEEEauthorblockN{Jani Boutellier}
\IEEEauthorblockA{University of Vaasa\\
Vaasa, Finland\\
jani.boutellier@univaasa.fi}
\and 
\IEEEauthorblockN{Juho Kannala}
\IEEEauthorblockA{Aalto University\\
Espoo, Finland\\
juho.kannala@aalto.fi}
}

%


\maketitle

\begin{abstract}
Camera pose estimation in large-scale environments is still an open question and, despite recent promising results, it may still fail in some situations. The research so far has focused on improving subcomponents of estimation pipelines, to achieve more accurate poses. However, there is no guarantee for the result to be correct, even though the correctness of pose estimation is critically important in several visual localization applications, such as in autonomous navigation.
In this paper we bring to attention a novel research question, pose confidence estimation, where we aim at quantifying how reliable the visually estimated pose is.
We develop a novel confidence measure to fulfill this task and show that it can be flexibly applied to different datasets, indoor or outdoor, and for various visual localization pipelines.
We also show that the proposed techniques can be used to accomplish a secondary goal: improving the accuracy of existing pose estimation pipelines.
Finally, the proposed approach is computationally light-weight and adds only a negligible increase to the computational effort of pose estimation. 
\end{abstract}


%
\IEEEpeerreviewmaketitle

\section{Introduction}
\label{ch:intro}
Visual localization aims at estimating the camera pose, i.e. position and orientation, from a given picture, under the assumptions that the camera intrinsic parameters and the environment in which the picture was taken are known. Particularly, the latter implies that a 3D model of the environment and/or a large database of images with annotated 3D coordinates are available. The interest for visual localization stems from its wide range of potential applications, from augmented reality to egomotion tracking. 

In robotics, autonomous vehicles need to keep track of the path they have travelled. This tracking problem is often approached using visual-inertial odometry.
 Unfortunately, all tracking methods are prone to error accumulation and the position estimate will eventually drift. 
 To overcome this problem, a reliable external signal, measuring the position of the device, must be used to correct the accumulated error. In outdoor environment, this task can be accomplished by GNSS/GPS \cite{crassidis2006}. However, as GNSS signals are not detectable inside buildings, other solutions are needed for indoor navigation. Together with RF-based (e.g. WiFi) positioning \cite{yang2015, dortz2012}, visual localization offers appealing possibilities \cite{bloesch2015, usenko2016}, and some state-of-the-art algorithms can already be run in real-time on mobile devices \cite{lim2014, lim2012}. As another application example, Augmented Reality (AR) also requires information about the camera pose \cite{marchard2016}. As objects are perceived differently in shape and size depending on the point of view of the observer, adding realistic-looking objects to the observer's view requires precise information about the camera's position and orientation. 
\begin{figure}[tb]
    \centering
    \begin{subfigure}{0.48\columnwidth}
    \includegraphics[width=\columnwidth]{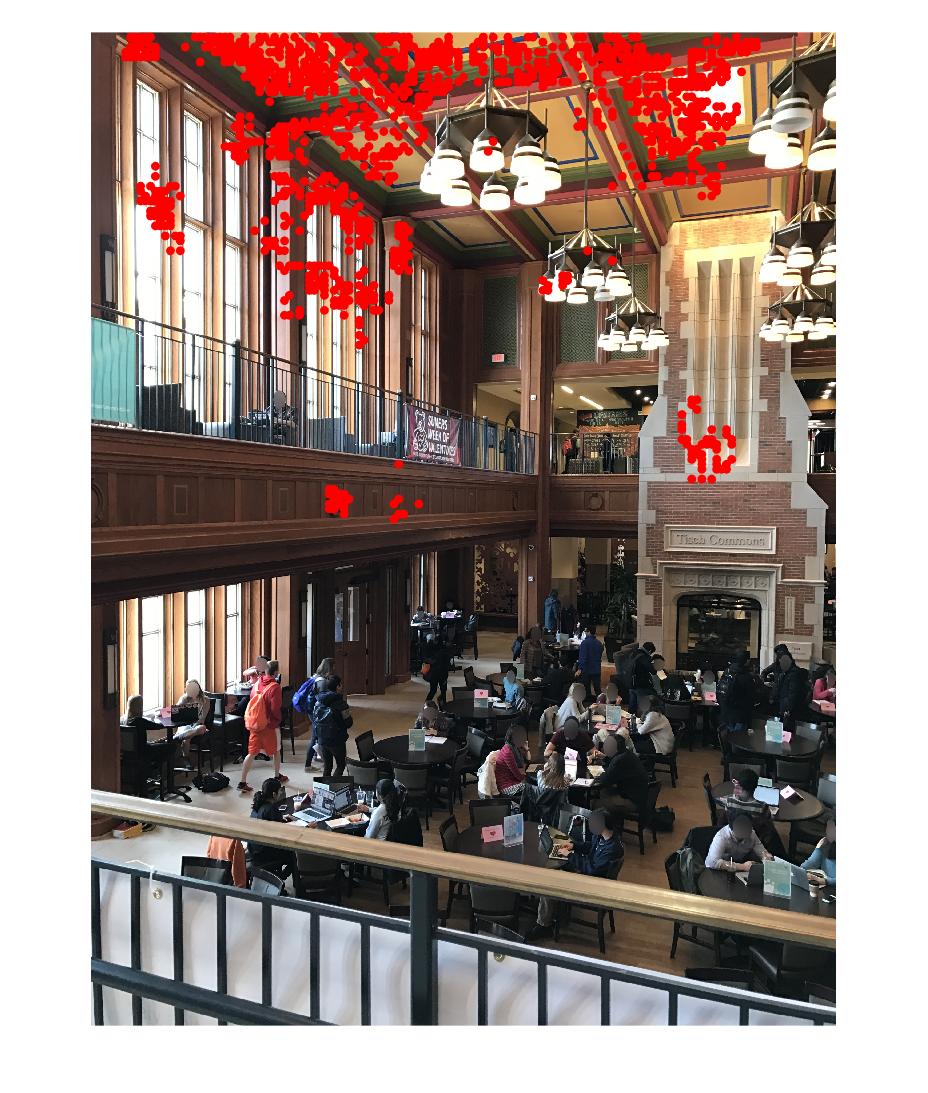}
    \end{subfigure}
    \begin{subfigure}{0.48\columnwidth}
    \includegraphics[width=\columnwidth]{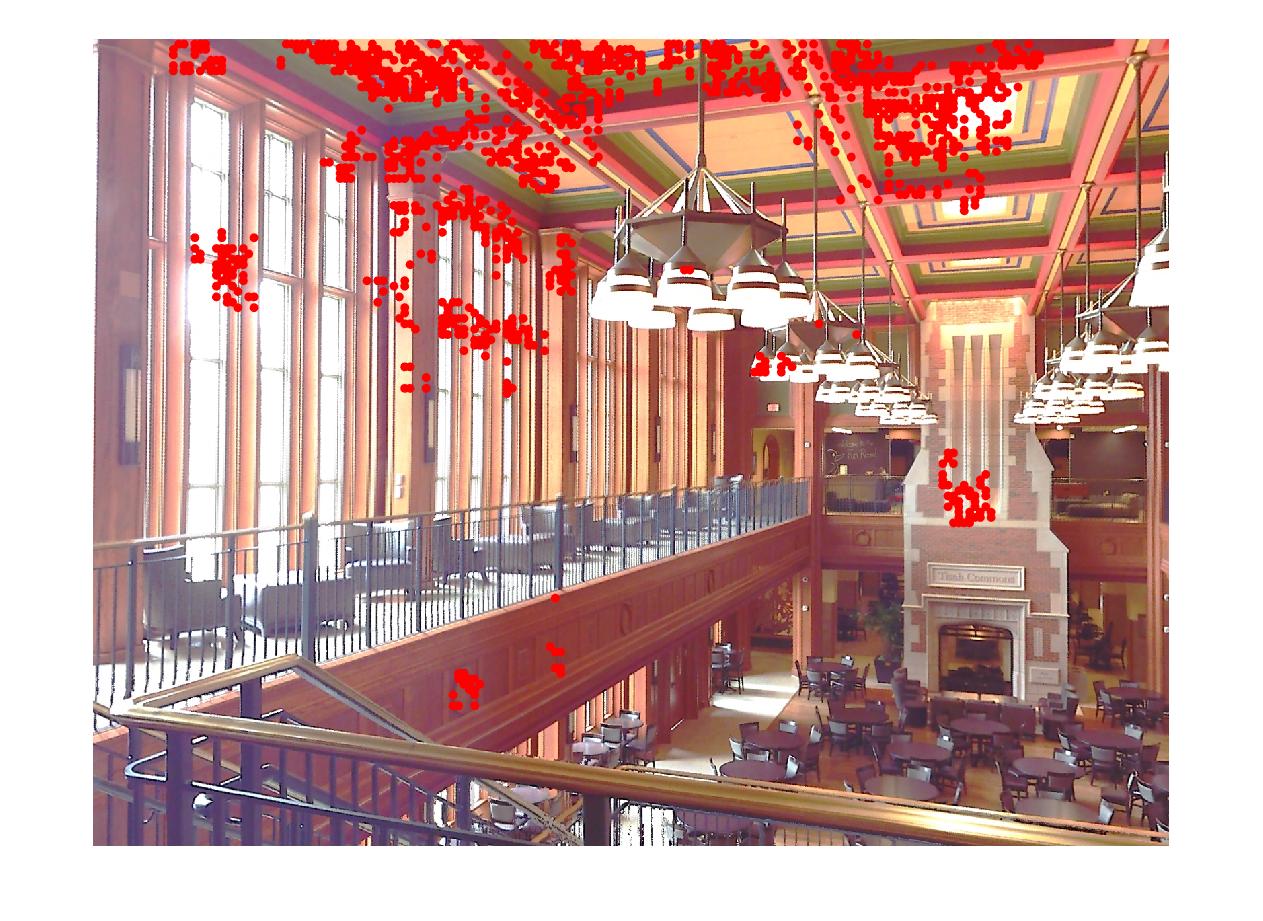}
    \end{subfigure}
    \caption{The query image (left) and the retrieved database image (right). Despite having 1247 inliers, the estimated position of the camera is 13.6 m apart from the corresponding ground truth.}
    \label{fig:inlsFails}
\end{figure}

\subsection{Motivation}
Given a \textit{query image} to be localized, most visual localization approaches form 2D-3D correspondences between pixels in the query image and points in the 3D space. These correspondences can be formed either directly, or by forming 2D-2D correspondences between the query image and an image whose 3D coordinates are known for most pixels. The camera pose can then be computed from these correspondences. The computation is generally wrapped within a RANSAC \cite{fischler1981} loop, where several hypotheses are generated and as a final result, the hypothesis with most correspondences is chosen as the final result. Those correspondences supporting the final hypothesis are referred as \textit{inliers}. 

The number of inliers has been so far the standard metric to assess the confidence of the estimated pose, i.e. the more inliers, the higher the probability for the hypothesis to be correct. Nevertheless, the number of inliers may not be robust enough in several situations. Figure \ref{fig:inlsFails} shows an example from InLoc, a retrieval based pipeline for indoor localization. The figure shows a query image (left) and a retrieved database image (right). In the Figure \ref{fig:inlsFails} case, the final result of the RANSAC loop reports 1247 inliers, however the estimated camera pose is still 13.6 m away from the ground truth. At first sight, it may appear that the presented image pair should lead to an accurate estimate, as the inliers seemingly are in the same places in both the query and in the database image. However, in reality, most of the inliers are located on the ceiling, where a strongly repeating pattern is present. As a consequence, despite inliers can be found in the same image areas, the inliers are matched incorrectly between the clusters. Repeating patterns together with occlusions, lack of texture, and variations in illumination are some of the main causes that make visual localization challenging and lead to failure despite a high number of inliers. Nevertheless, in many practical applications is crucial to know which pose estimates provided by visual localization can be trusted and that is the problem we aim to address in this work. 

As we will discuss in Section 2, also other confidence measures than the count of inliers have been proposed. The work so far, however, takes place \textit{inside} pose estimation pipelines and aims at achieving more accurate pose estimates. Despite the improvements in the last years \cite{sarlin2019, taira2018, sattler2012}, pose estimation is still far from being solved and, despite all developments to pose estimation pipelines, the final result may still be wrong. As failure is something that just needs to be coped with, we present in this paper a new research question, referred as \textit{confidence estimation}. Opposed to the previous work, our confidence estimator is placed \textit{after} the pose estimation pipeline and aims at quantifying how reliable the final result is. The presented research question is more general than the ones addressed so far. In our treatment, we develop a more general metric, which can be used to compare poses from different query images, whereas previous work is limited to considering one query image at a time. The final goal is to increase the self-awareness of algorithms regarding their own limits, in order to critically judge whether the final result should be accepted or rejected. Introducing this additional confidence estimation step after the pipeline can be beneficial e.g. in odometry, where several query images are available from a video recording of the environment. After the final pose for each frame has been computed, our confidence measure can be used to choose which frames lead to a reliable pose estimate.

\subsection{Contribution and Structure of the Paper}
The contribution of this paper can be summarized as follows:
\begin{itemize}
    \item We formulate the pose confidence estimation problem, aiming at quantifying how reliable the final pose is. Opposed to previous work, our method takes place after the visual localization pipelines and it can be applied to compare the confidence of several query images.
    \item We propose a novel \textit{normalized confidence score}, more robust than the inliers count. We show that our confidence estimation algorithm is independent of the dataset and error threshold used for training, and can be applied to several pipelines and datasets.
    \item We show how our method, in addition to giving a confidence measure of the final result, can also be integrated inside existing visual localization pipelines to achieve more accurate poses.
\end{itemize}

This paper is structured as follows: in Section~2 the related work is presented. In Section~3 we analyze in more details \textit{InLoc}, our main case study pipeline. In Section~4 we present a formalization of the confidence estimation problem and describe how our confidence measure is computed. In Section~5 we give technical details of the training process. In Section~6 we present the results of our experiments, mainly focusing on how our algorithm perform on the InLoc dataset, but also showing that it generalizes to other datasets. The results are finally summarized in Section~7.

\section{Related Work}
\label{ch:relatedWork}
In this section, we briefly review relevant related work. Particularly, we focus on what are the state-of-the-art approaches for visual localization, emphasizing their strengths and weaknesses. We also review how the confidence estimation problem and uncertainty quantification has been tackled for visual localization applications.
\subsection{Visual Localization}
Visual localization aims at computing the camera pose $P$ from a given image. Mathematically, this is equivalent to computing the euclidean transformation $[R\quad\vec{t}]$ from the real world to the camera coordinate system, where $R$ and $\vec{t}$ are the rotation matrix and the translation vector, respectively.  

\textbf{Direct matching approaches} establish 2D-3D point correspondences directly between the query image and the 3D model \cite{sattler2011, sattler2012}. Next, the camera pose can be estimated using a Perspective-n-Points (PnP) solver \cite{kneip2011, xie2012, gao2003} within a RANSAC loop \cite{fischler1981}. These approaches allow very accurate pose estimation but are generally time consuming. 

\textbf{Image retrieval-based approaches:} An appealing compromise between efficiency and accuracy is offered by retrieval based pipelines \cite{sarlin2019, taira2018}, where the knowledge of the environment is expressed by a large database of images, for which the 3D coordinates are known at the level of pixels. Given a query image, the most similar images are retrieved from the database and then 2D-2D correspondences are formed between query and database images. As the 3D coordinates of the database images are known, the 2D-2D correspondences induce 2D-3D correspondences, used to compute the camera pose with a PnP-RANSAC solver.

\textbf{Regression-based approaches} exploit deep neural networks to directly regress the pose from a given image \cite{kendall2015, melekhov2017}, or regress the 3D coordinates \cite{shotton2013, Brachmann_2017_CVPR, Brachmann_2018_CVPR, li2018} from the query image and compute the pose from the so obtained correspondences. Despite being computationally efficient, deep-learning based regression approaches have up to now led to lower accuracies than retrieval based or direct matching approaches \cite{sattler2019}. 

\subsection{Uncertainty in Visual Localization}
\textbf{Confidence Estimation in Pose Estimation:} Kendall \textit{et al.} proposed in \cite{kendall2016} a Bayesian pose regression neural network which, instead of producing a point prediction of the pose, produced a confidence interval for each scalar parameter. This was included into their previously developed framework \cite{kendall2015}, allowing to improve the accuracy of pose estimation. The work was based on \cite{gal2015}, where it was shown that \textit{dropout} can be exploited to transform a neural network into a Bayesian neural network. This approach is suitable for pose regression approaches but is not applicable to other approaches of visual localization. In \cite{taira2018} a set of candidate poses is generated for each query image and a novel metric, obtained by pose verification (PV), is used to choose the best candidate, outperforming the number of inliers measure. In \cite{taira2019}, semantic segmentation is exploited to determine the best candidate pose for a single query image.

\textbf{Confidence Estimation in RANSAC:} RANSAC is central in most visual localization pipelines and it is employed both while forming point correspondences and while computing the camera pose. In its standard form, the hypothesis with the highest number of inliers is regarded as the best one. Some variants \cite{chum2005, raguram2009} first estimate the uncertainty in the input points to randomly sample only from the most confident ones. In \cite{torr2000}, an alternative approach, maximizing the likelihood instead of the  number of inliers is proposed.  In \cite{Hassner2013} RANSAC for forming 2D-2D points correspondences is analyzed and an alternative metric, obtained by weighting several parameters such as inliers counts, their spatial distribution and similarity of images, was proposed to form more robust point correspondences.

Alternatives to the number of inliers measure described above address intermediate steps of visual localization pipelines to increase the accuracy of the final estimate. Thus they are all computed for a single query image. In this paper, we address a wider problem. We develop a more general metric, which can be used to compare the confidence of different query images. The main purpose is not to achieve a more accurate pose, but to be able to automatically grade how reliable the final pose is. On the other hand, choosing the best candidate pose for a single query image, as is generally done in retrieval approaches, can be regarded as a particular case of our problem, which our method can also solve more robustly than the inliers count alone.

\section{Case Study}
\label{ch:inloc}
\begin{figure}[tb]
	\centering
	\includegraphics[width=\columnwidth]{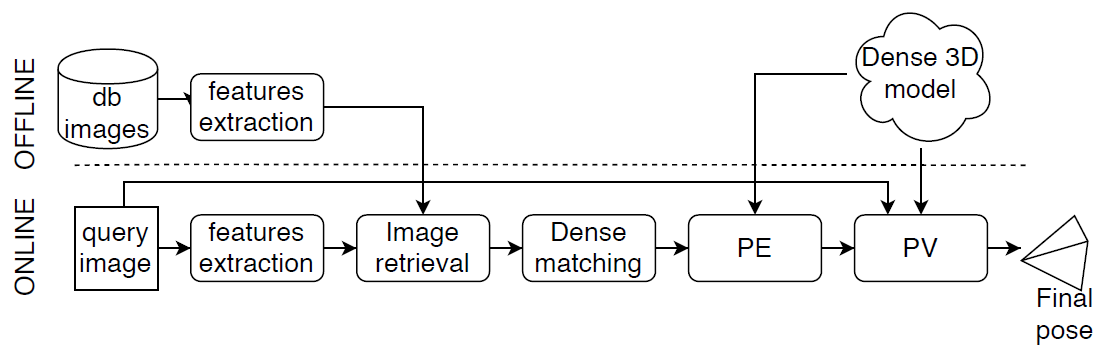}
	\caption{The InLoc visual localization pipeline. PE and PV refer to Pose Estimation and Pose Verification, respectively.}\label{fig:inloc}
\end{figure}

As a case study for our confidence estimation problem, the indoor visual localization algorithm InLoc \cite{taira2018} is used. 
The structure of the InLoc pipeline is depicted in Figure \ref{fig:inloc}.

InLoc is a retrieval-based pipeline, where global features for image retrieval are extracted with the NetVLAD \cite{arandjelovic2016netvlad} convolutional neural network. Point matching is done in a coarse-to-fine manner, where local features are extracted using the VGG-16 network \cite{simonyan2014very}. The camera pose is then estimated using P3P-LO-RANSAC \cite{lebeda2012fixing}. At the end of the pipeline, the authors introduce a novel pose verification (PV) step, which is used to choose the best candidate pose. In this phase, each estimated pose is used to synthesize a new picture from a dense 3D model of the environment. The synthesized views are then compared to the original query image and a \textit{verification score}, measuring their similarity is computed. The final result of the pipeline is the candidate pose with the highest verification score. 

The authors have also developed a dataset of the same name to benchmark their algorithm \cite{wiijmans2017}. The dataset presents several indoor scenes taken from two different buildings on different floors of Washington University. The pictures of the dataset were collected during different periods, producing several challenges for pose estimation pipelines, such as differences in illumination and view points, as well as varying occlusions. The dataset contains a database of 9972 images, for which the original 3D coordinates of most pixels are known, and 329 query images used to test the algorithm. In their work, the authors reach an accuracy of ~70\% using $\SI{1}{\meter}$, $\SI{10}{\degree}$ error threshold. Due to its elevated degree of challenge, this dataset was chosen as the main case study for our confidence estimation problem.

\section{Confidence Estimation}
\label{ch:confEst}
Given an estimated pose $\hat{P}$, we want to map it to a scalar $\gamma\in[0,1]$, to quantify the reliability of $\hat{P}$. If an error threshold for the pose is fixed, $\gamma$ can be used to classify the estimate as correct or incorrect, by choosing a threshold $\gamma_0$ and accepting only the poses with $\gamma\geq\gamma_0$. The choice of $\gamma_0$, however, depends on the particular application and on the predetermined error threshold. A good confidence measure, however, should be able to rank the computed poses based on their confidence regardless of the chosen threshold. The challenge is now to find a function $f:\hat{P}\mapsto
\gamma$, to compute the confidence score of the estimated pose. In Sections 4.1 and 4.2, we describe what features are extracted from $\hat{P}$ to model its confidence and in Section 4.3 the final algorithm, for computing $\gamma$ from the features is presented.

\subsection{Number of Inliers}
\begin{figure}[tb]
    \centering
    \includegraphics[width=0.48\columnwidth]{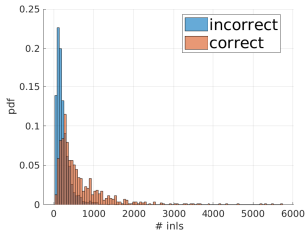}
    \caption{Distribution of inliers for correct and incorrect poses. Despite correct poses having on average a higher number of inliers, the histograms still overlap significantly, making the inliers count alone unsuitable for discriminating the classes.}
    \label{fig:inlsDist}
\end{figure}

In RANSAC-based pipelines the number of inliers of the final pose is the standard metric to quantify its confidence. In challenging environments, however, this alone is not always reliable. Particularly, in the context of indoor localization, where repeating textural patterns are common, an inaccurate pose may be obtained despite the high number of inliers.
This is observed in Figure \ref{fig:inlsDist}, where the inliers distributions for correct and incorrect poses of InLoc are plotted. Here, the pose is regarded as correct if the translational error is below $\SI{1}{\meter}$ and the angular error below $\SI{10}{\degree}$. As can be noticed, the histogram of correctly localized poses has indeed a longer tail, indicating that estimates with more than 2000 inliers are likely to be correct. However, these cases are relatively rare, as both histograms have most of the mass below 1000, where they are significantly overlapping. As a conclusion, the number of inliers alone cannot be used to distinguish between successful and unsuccessful estimates. For this reason, in the two next subsections we investigate other factors that need to be taken into account and in Subsection 4.3 we finally discuss how to combine the different observations. 

\subsection{Inliers Spatial Distribution}
\begin{figure}[tb]
    \centering
    \begin{subfigure}{0.48\columnwidth}
    \includegraphics[width=\columnwidth]{figures/introduction/queryInls.jpg}
    \end{subfigure}
    \begin{subfigure}{0.48\columnwidth}
    \includegraphics[width=\columnwidth]{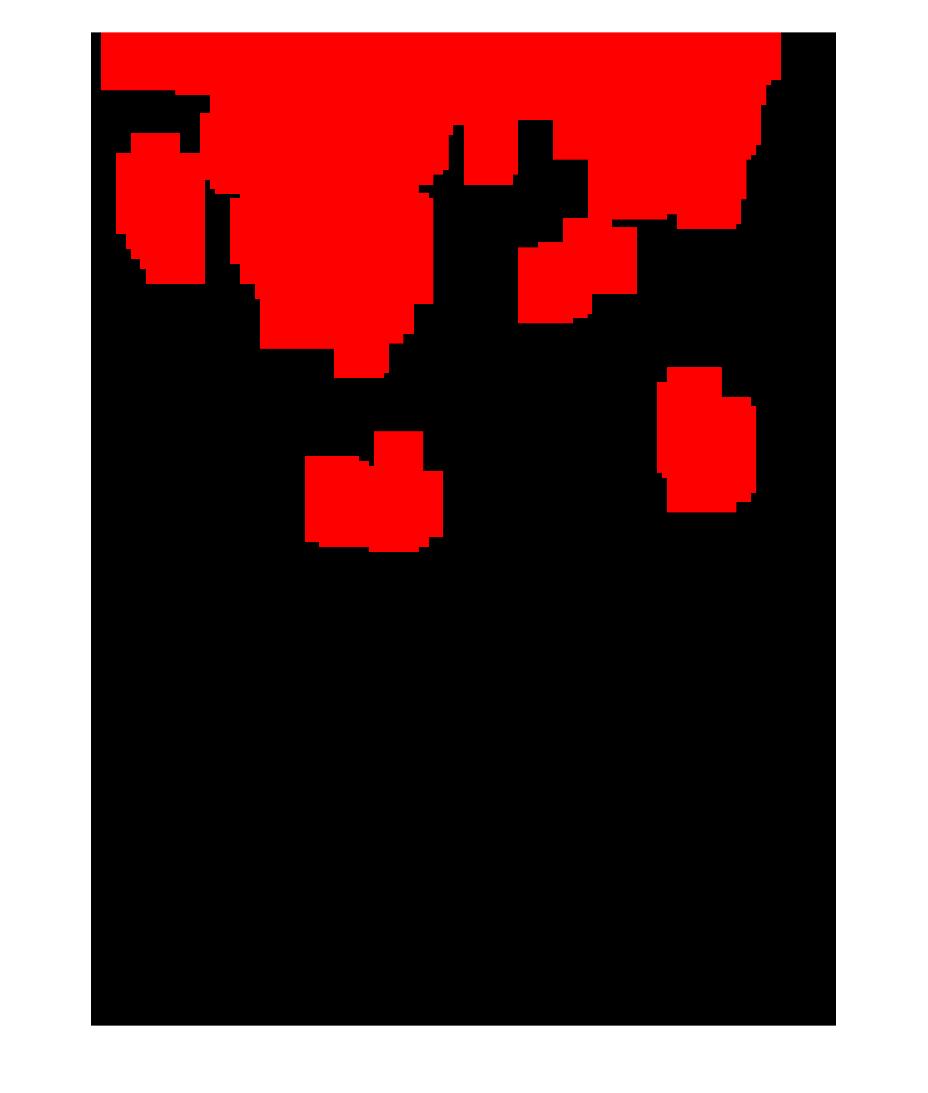}
    \end{subfigure}
    \caption{Query image with annotated inliers (left) and corresponding coverage map (right).}
    \label{fig:queryCov}
\end{figure}

\begin{figure}[tb]
    \centering
    \begin{subfigure}{0.48\columnwidth}
    \includegraphics[width=\textwidth]{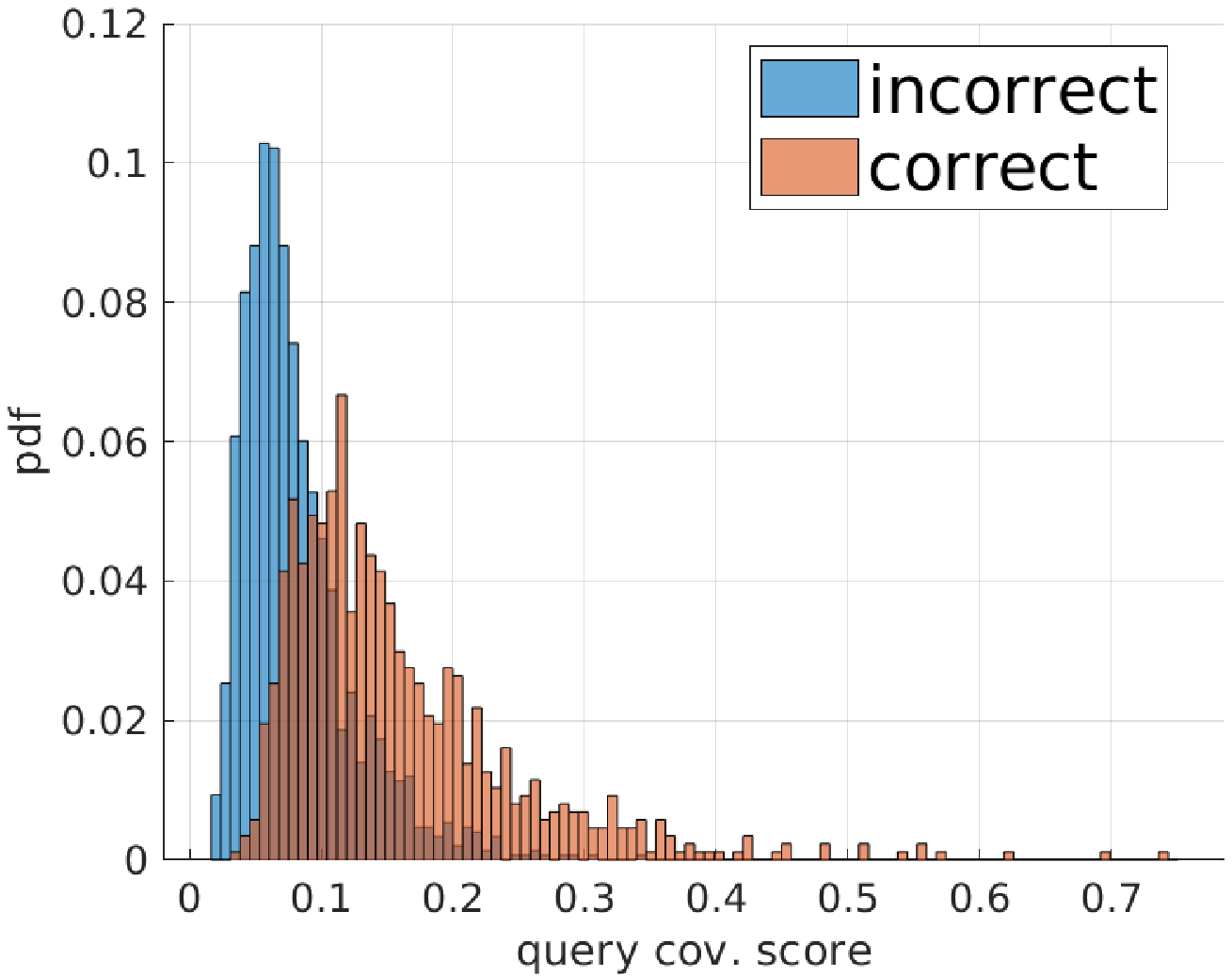}
    \end{subfigure}
    \begin{subfigure}{0.48\columnwidth}
    \includegraphics[width=\textwidth]{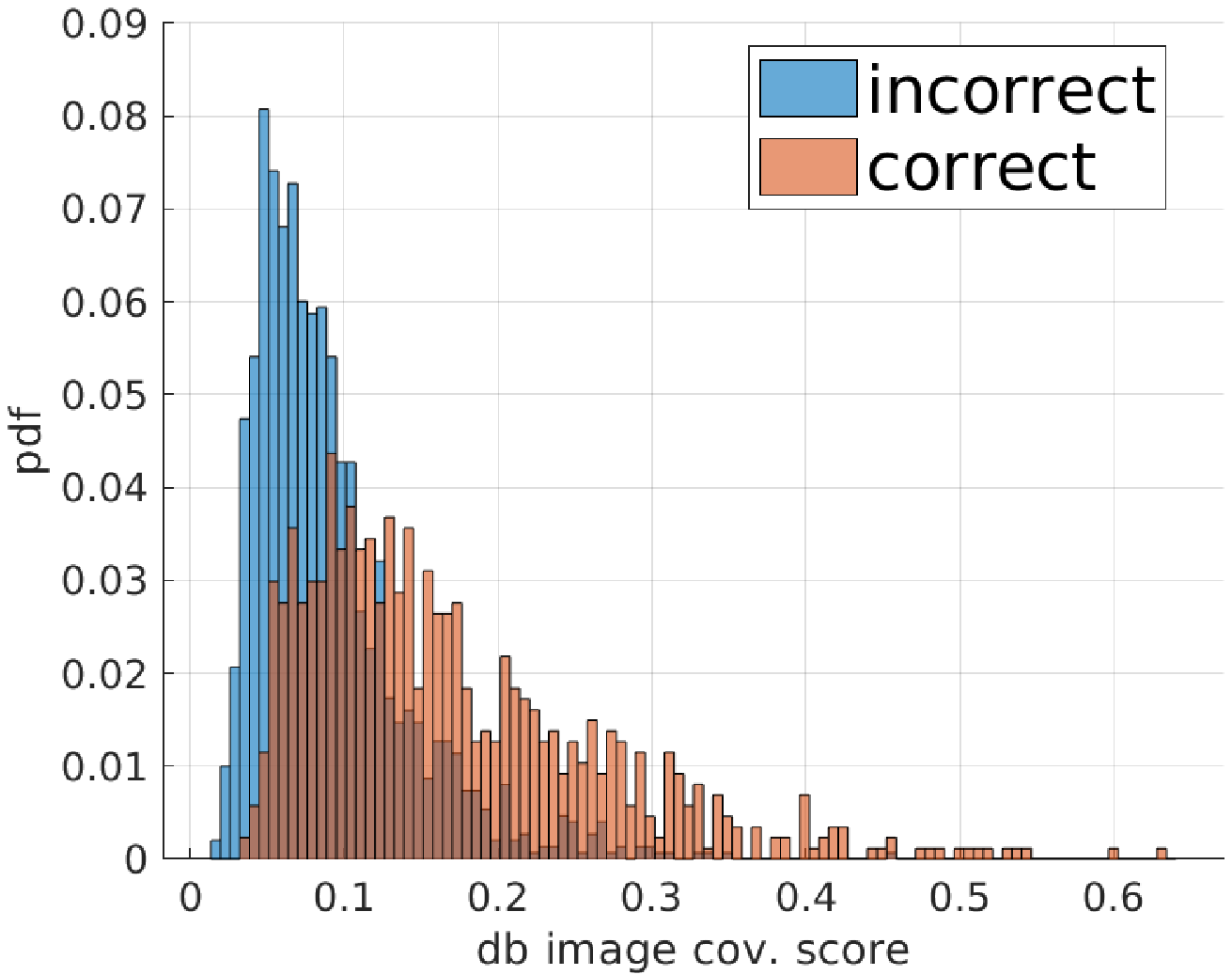}
    \end{subfigure}
    \caption{Coverage scores distributions for correct and incorrect poses.}
    \label{fig:covHist}
\end{figure}

Due to redundant patterns in images, the number of inliers alone is not a good metric to assess the confidence of the final result. Particularly, if all the inliers are condensed into a small area of the picture, it is more likely that they indicate the presence of a similar-looking object in both pictures, even if in completely different environments. For this reason, we take into account also the spatial distribution of the inliers, penalizing those cases where the inliers cover only a small area. This was already tried in \cite{Hassner2013}, where the covered area of the inliers was computed as a convex hull. The convex hull, however, may still be significantly bigger than the area covered by inliers, especially if they form several small clusters, such as in Figure \ref{fig:queryCov}. To quantify the covered area here, we propose a different more robust approach. We say a pixel is \textit{covered} if there is at least one inlier in its neighbourhood and our final \textit{coverage score} is the number of covered pixels divided by the total number of pixels in the image. This is visualized in Figure \ref{fig:queryCov}, where a \textit{coverage map}, highlighting the covered inliers is shown. In this approach, the size of the neighbourhood is the only parameter to be tuned. A too big neighbourhood would give a too optimistic coverage score and a too small neighbourhood would be equivalent to counting inliers. In this work, we chose to use a rectangular neighbourhood, whose sizes are 1/15 of the size of the image. The size of the neighbourhood was chosen by visual inspection of the obtained coverage maps. The parameter could have also been optimized by grid search, however numerical experiments revealed that small variations in the neighbourhood size did not affect significantly the final confidence score computed in Section 4.3.
Figure \ref{fig:covHist} shows the histograms of coverage scores for successful and unsuccessful poses. Again, overlap between histograms is still present and coverage scores alone cannot thus be used as a reliable confidence measure.

\subsection{Learning to Predict a Confidence Measure}
As shown in the previous sections, none of the parameters alone allows to discriminate linearly between successful and unsuccessful estimates, as the histograms overlap significantly. Nevertheless, each parameter described above correlates with the success of the pose. The next task is to combine the information obtained from different scalar parameters into the final confidence measure $\gamma$. As the number of input parameters is low and we want our approach to work generally on different datasets and error thresholds, we chose to use the logistic regression algorithm to compute the final confidence score. Let $x_1$, $x_2$ $x_3$ the number of inliers, query image coverage score and database image coverage score, respectively. The final confidence score will then be
\begin{equation}
    \gamma=\textrm{logsig}\left(b+\sum_{i=1}^3w_ix_i\right), \label{eq:logistic}
\end{equation}
where $\mathrm{logsig}(x)=\frac{1}{1+e^{-x}}$ is the logistic function and $w_1$, $w_2$, $w_3$ and $b$ are the parameters to be learned.

Using a simple algorithm reduces the risk of overfitting to the properties of the dataset used for training. Furthermore, our confidence estimator is an additional task for the visual localization pipeline, as it takes place after the final result is computed. This could be problematic in real-time applications, however using our tiny algorithm, the additional computation time is negligible.

\section{Model Training with InLoc}
\label{ch:experiments}
Our confidence estimation algorithm can be applied \textit{after} the camera pose has been estimated, meaning the 329 InLoc query images have to be used to both train and test our method.
To overcome the problems stemming from the small amount of training data, we produce an \textit{extended dataset} as follows: for each query image, all 10 candidate database images provided by the InLoc pipeline are taken into account, obtaining 3290 query and database image pairs, each associated with an estimated camera pose. Out of these, we remove those pairs that have less than 3 point correspondences, as they represent a trivial case of failed estimation. For each pair, the coverage scores can be computed from the inlier positions produced by RANSAC. Each estimated pose is labelled as correct (1) if the error is below the threshold of $\SI{1}{\meter},\SI{10}{\degree}$ and as incorrect (0) otherwise. As a result, our extended dataset presents 2368 estimated poses, out of which 75\% are used as training data and the rest as test data. The training-test splitting is done so that all the image pairs related to the same query image are either in the training set or test set, but not in both. Overall, our training dataset contains 1765 query-database images pairs, related to 230 distinct query images and the test dataset contains 603 query-database image pairs, corresponding to 99 distinct query images.

\section{Results}
\label{ch:results}
In this section we present the results of our quantitative experiments. To assess the performance of our confidence estimation algorithm, precision-recall curves are plotted and the Area Under Curve (AUC) is used as the evaluation metric. The main part of the results focuses on the InLoc dataset, however at the end we also investigate how our method generalizes to different datasets.

\subsection{Confidence Estimation on InLoc}
\begin{figure}[tb]
    \centering
    \begin{subfigure}{0.47\columnwidth}
    \includegraphics[width=\textwidth]{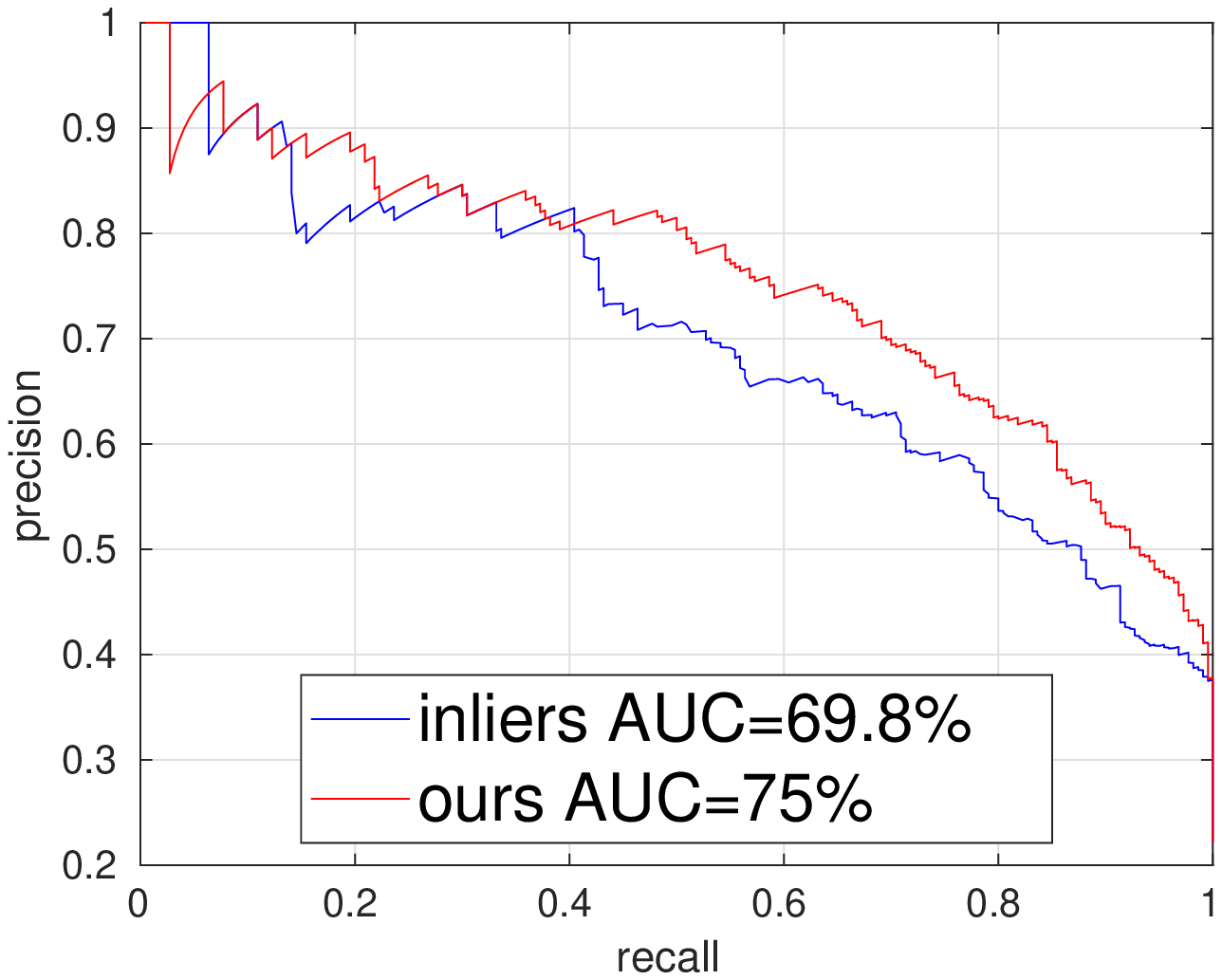}
    \caption{Extended dataset}
    \end{subfigure}
    \begin{subfigure}{0.47\columnwidth}
    \includegraphics[width=\textwidth]{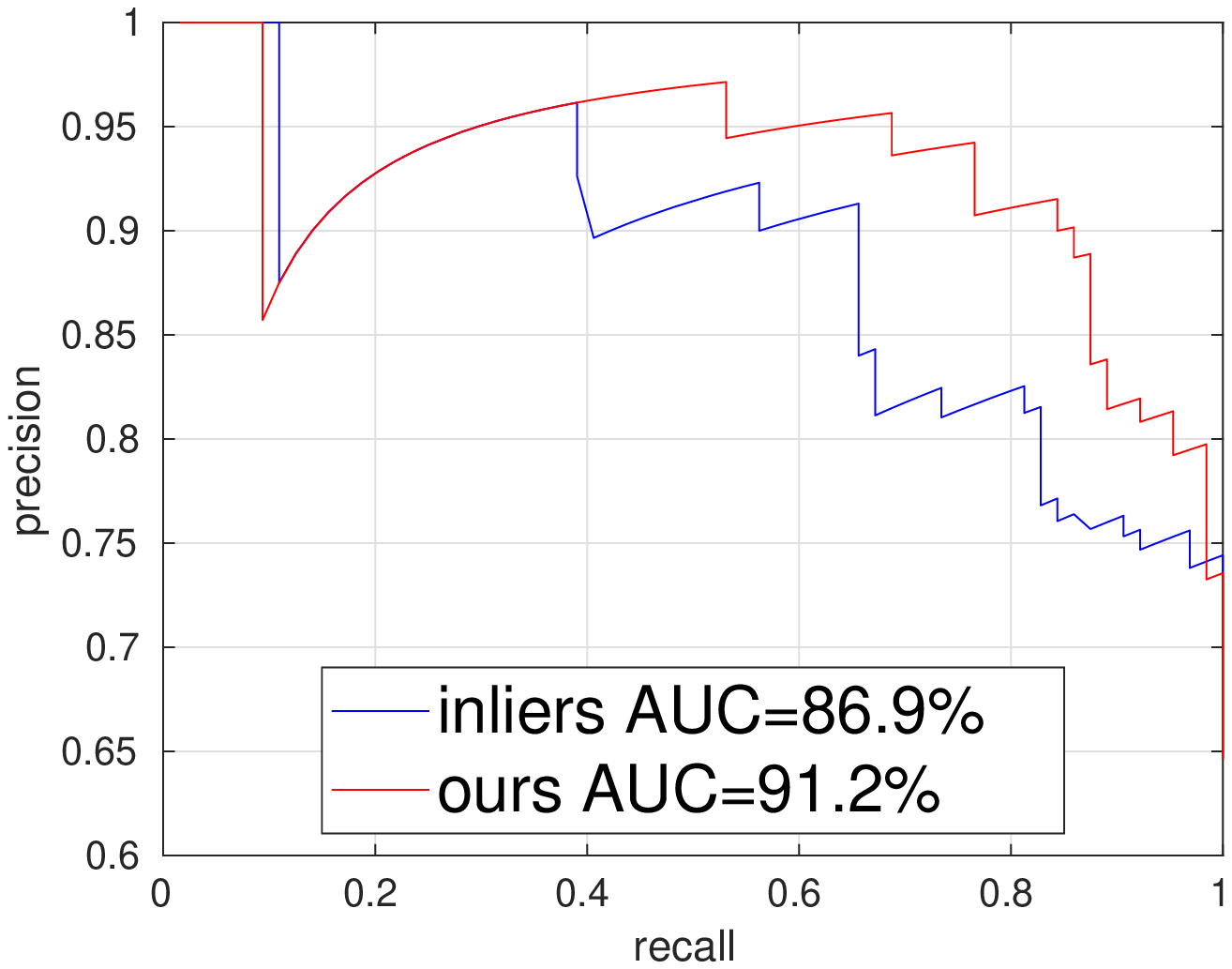}
    \caption{Only best candidates}
    \end{subfigure}
    \caption{Precision-recall curves when classifying test data as correct or incorrect.}
    \label{fig:prBasic}
\end{figure}

Figure \ref{fig:prBasic} shows the precision-recall curve (PR-curve) when classifying the test data of the extended dataset as correct or incorrect using the classical number of inliers (blue curve) and the confidence measure computed with our method (red curve). As can be noticed, our method is more robust than the inliers count alone, \textbf{improving the AUC from 69.8\% to 75\%.}

The extended dataset was obtained taking all ten candidates for each query image. Clearly, this introduces more incorrect poses than correct ones, making our dataset biased. In practical applications, however, most of these wrong poses are already filtered out inside the pipeline when picking the best candidate. For this reason, we also evaluate our method on a subset of the test data, which contains only the best candidate for each query image. This experiment is possible, because in the training-test splitting phase all estimated poses related to the same query image were included entirely either in the training or in the test set. As the right-hand graph in Figure \ref{fig:prBasic} shows, filtering out incorrect candidate poses does not affect the performance of our model, as the AUC still \textbf{improves from 86.9\% to 91.2\%.}

The main difference between our approach and the classical number-of-inliers-only lies in the fact that our model needs to be trained, whereas the inliers count is just a number, whose computation does not depend on the dataset or pipeline used. Particularly, our method was trained fixing the error threshold to 1~m and $10^\circ$. For this reason, in Table \ref{tab:diffThres} we report the AUCs obtained with our method at different thresholds \textit{without changing the threshold used to train it}. The results were obtained using only the best candidates for each query image. As Table \ref{tab:diffThres} shows, using a stricter threshold makes the classification problem harder in general, but the proposed model is still more robust than the baseline.

\begin{table}[tb]
    \centering
    
    \caption{AUCs with inliers count and our method at different error thresholds.}
    \label{tab:diffThres}
    \begin{tabular}{c|c|c}
         Error Threshold&AUC (inls count)&AUC (ours)  \\\hline\hline
         1.5~m, $10^\circ$&87.1\%&91.8\%\\
         1~m, $10^\circ$& 86.9\%&91.1\%\\ 
         0.5~m, $10^\circ$&68.7\%&76.8\%\\
         0.25~m, $10^\circ$&51.8\%&56.8\%\\
    \end{tabular}
\end{table}
\subsection{Ablation Study}
It is also interesting to investigate the contribution of each input parameter to the final confidence measure. For this reason, we perform an ablation study on our model, removing each parameter one at a time and reporting how the AUC on the test data changes. The results are shown in Table \ref{tab:ablation}.
\label{ss:ablation}
\begin{table}[tb]
    \centering
    \caption{Ablation study of the proposed model.}
    \label{tab:ablation}
    \begin{tabular}{c|c}
         Case&AUC\\\hline\hline
         Full model& 75.0\%\\
         Only inliers & 69.8\%\\
         Inliers and db image cov. score&69.4\%\\
         Inliers and query image cov. score&74.0\%\\
         Query and db image cov. score&72.3\%\\
    \end{tabular}
\end{table}
Table \ref{tab:ablation} shows that the spatial distribution of inliers carries indeed useful information for assessing whether the final pose is reliable or not. Particularly, the coverage score of the query image seems to be more relevant than the respective score for the database image.

\subsection{Pose Verification for Confidence Estimation}
\begin{figure}[tb]
    \centering
    \begin{subfigure}{0.47\columnwidth}
    \includegraphics[width=\textwidth]{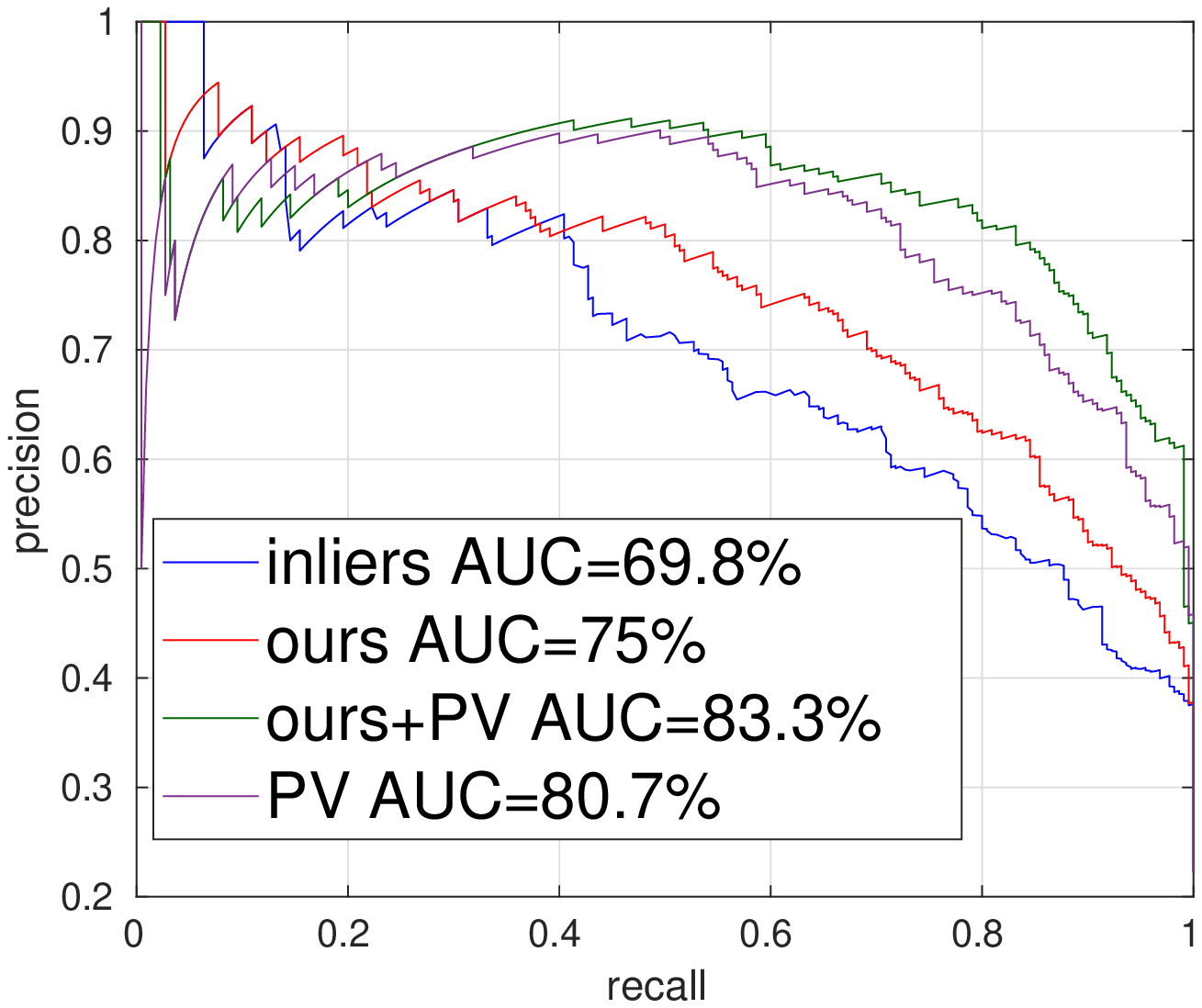}
    \end{subfigure}
    \begin{subfigure}{0.47\columnwidth}
    \includegraphics[width=\textwidth]{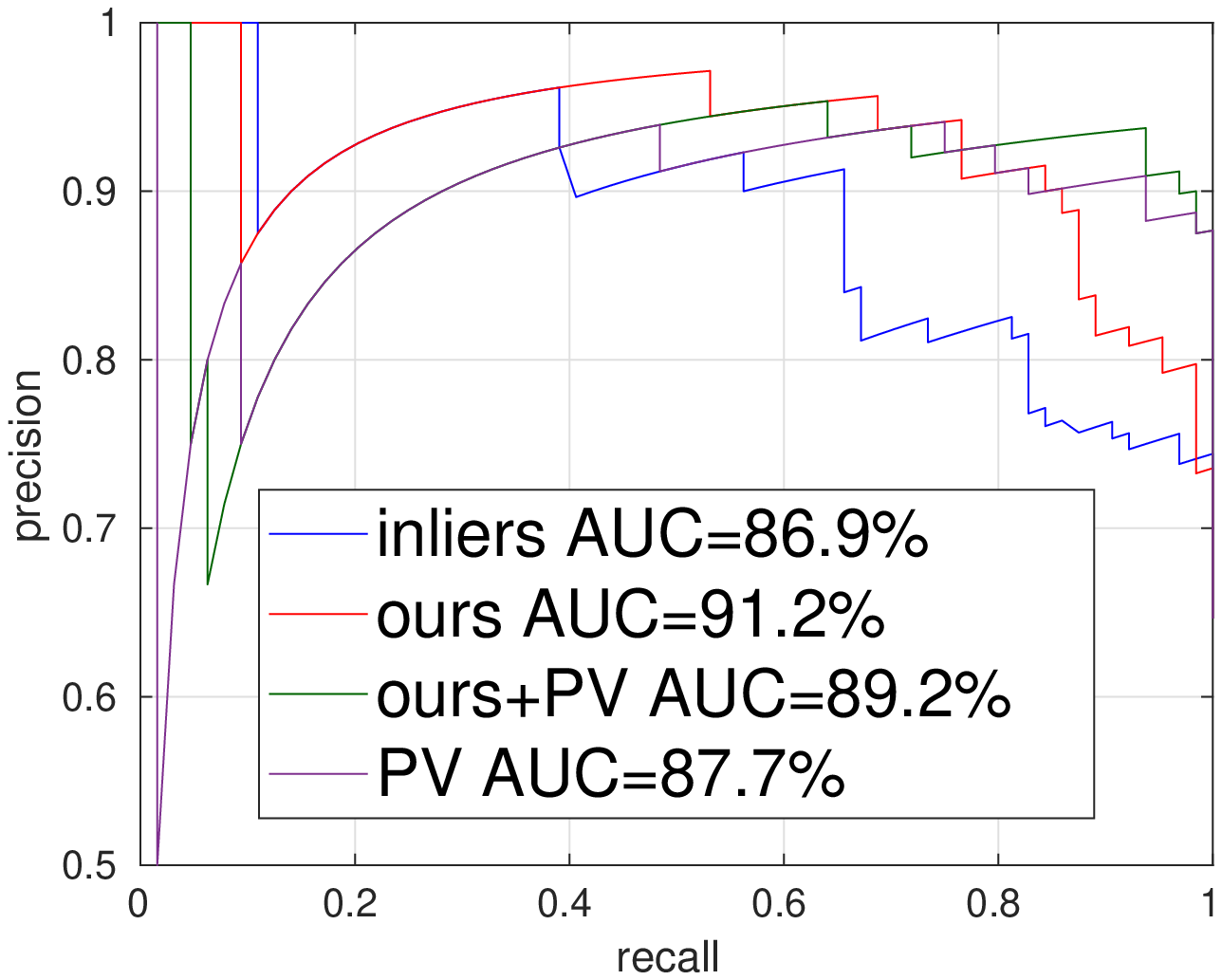}
    \end{subfigure}
    \caption{PR-curves when using the verification score as additional input to our model.}
    \label{fig:verificationInPR}
\end{figure}

In \cite{taira2018}, the authors of InLoc showed how pose verification improved the accuracy of the pipeline. It is hence reasonable to see whether it can be included into our confidence estimation model as well. To do so, we introduce InLoc verification score as a further parameter to Eq. \eqref{eq:logistic} and retrain the model as described above. Figure \ref{fig:verificationInPR} shows the PR-curves using the number of inliers (blue curve), our metric computed without verification score (red curve) and our metric computed with verification score (green curve). Interestingly, the behaviour is now different between the extended dataset and its subset with only the best candidate of each query. 
When testing with the extended dataset, the inclusion of the verification score seems to improve the AUC ~8\%. However, when testing only on the best candidate for each query image, our model without verification score performs the best, with our model with verification score still being more robust than the inliers count alone. As a lesson learnt, pose verification is useful inside the pipeline, when comparing poses related to the same query image. However, the verification score does not generalize well for scoring the confidences of different query images.

\subsection{Confidence Estimation for Higher Accuracy}
The output of our model $\gamma$ can be interpreted as the probability for the estimated pose to be correct. It has been emphasized throughout this work that the confidence estimation problem is more general than choosing the best candidate pose, because different query images are considered. On the other hand, if our algorithm works for several different query images, it should work also for the special case of a single query image. This motivates us to investigate whether our algorithm can be included inside the pose estimation pipeline to achieve better accuracies.
\begin{figure}[tb]
    \centering
    \includegraphics[width=0.47\columnwidth]{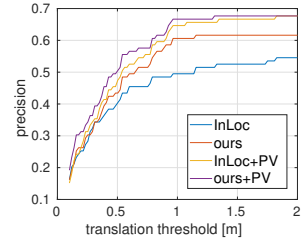}
    \caption{Integration of our confidence estimator into the InLoc pipeline. The angular error threshold is kept at $\SI{10}{\degree}$. Our method achieves better accuracies independently of the translation error threshold.}
    \label{fig:accuracies}
\end{figure}
\begin{table}[tb]
    \centering
    \caption{Accuracies with fixed error threshold $\SI{1}{\meter}$, $\SI{10}{\degree}$, showing our method outperforms the baseline InLoc, both with and without pose verification.}
    \label{tab:accuracies}
    \begin{tabular}{c|c|c}
         &InLoc&Ours\\\hline\hline
         Without pose verification&49.5\%&60.6\%\\
         With pose verification&64.7\%&66.7\%
    \end{tabular}
\end{table}
Table \ref{tab:accuracies} shows the obtained accuracies when ranking the candidate poses with the InLoc pipeline (number of inliers or verification score) and with our method (with and without InLoc verification score). Poses are considered correct using the $\SI{1}{\meter}$, $\SI{10}{\degree}$ threshold. For a fair comparison, only the 99 query images that were not used to train our model are considered, which leads to different accuracy values with respect to the original InLoc work. Our method is shown to improve the accuracy over the baseline, both with and without verification score, indicating that the proposed approach can also be included within visual localization pipelines to obtain more accurate poses. Furthermore, our model increases the accuracy regardless of the error thresholds, as Figure \ref{fig:accuracies} shows.

\begin{figure*}[t!]
    \centering
   
    \begin{subfigure}[t]{0.48\textwidth}
    \centering
        \includegraphics[height=3cm]{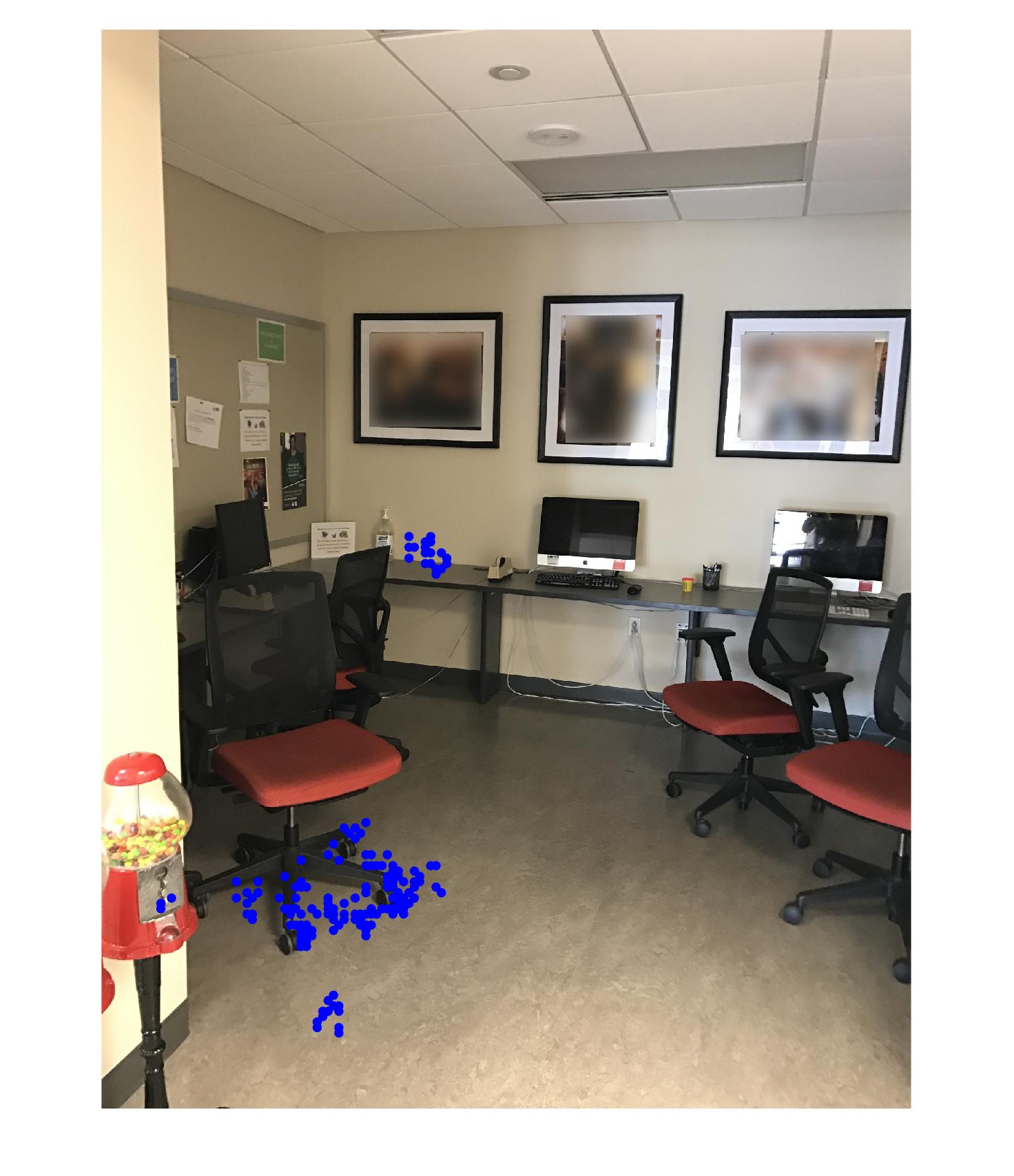}
        \includegraphics[height=3cm]{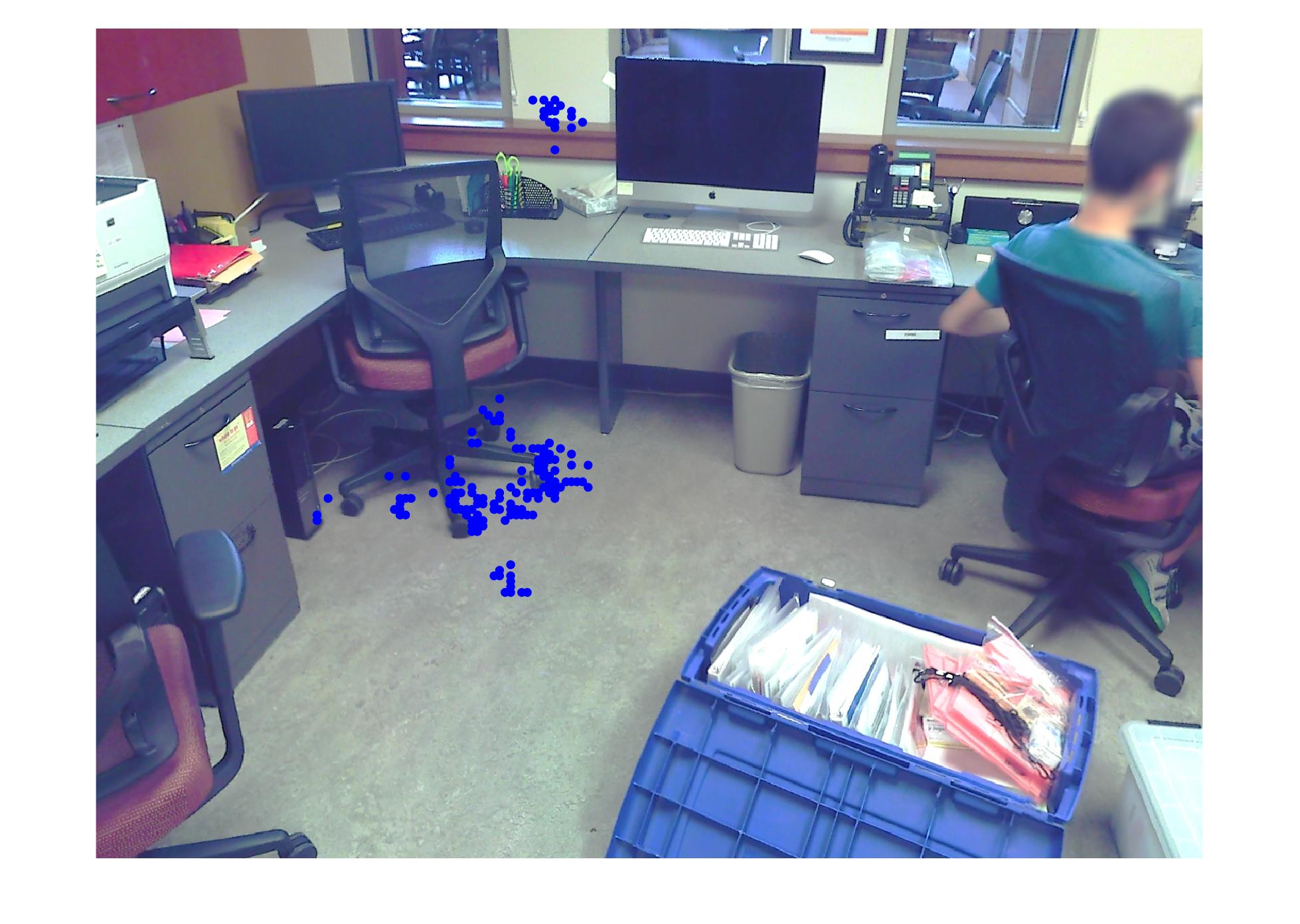}
    \end{subfigure}\hspace{0.1em}
     \begin{subfigure}[t]{0.48\textwidth}
        \centering
        \includegraphics[height=3cm]{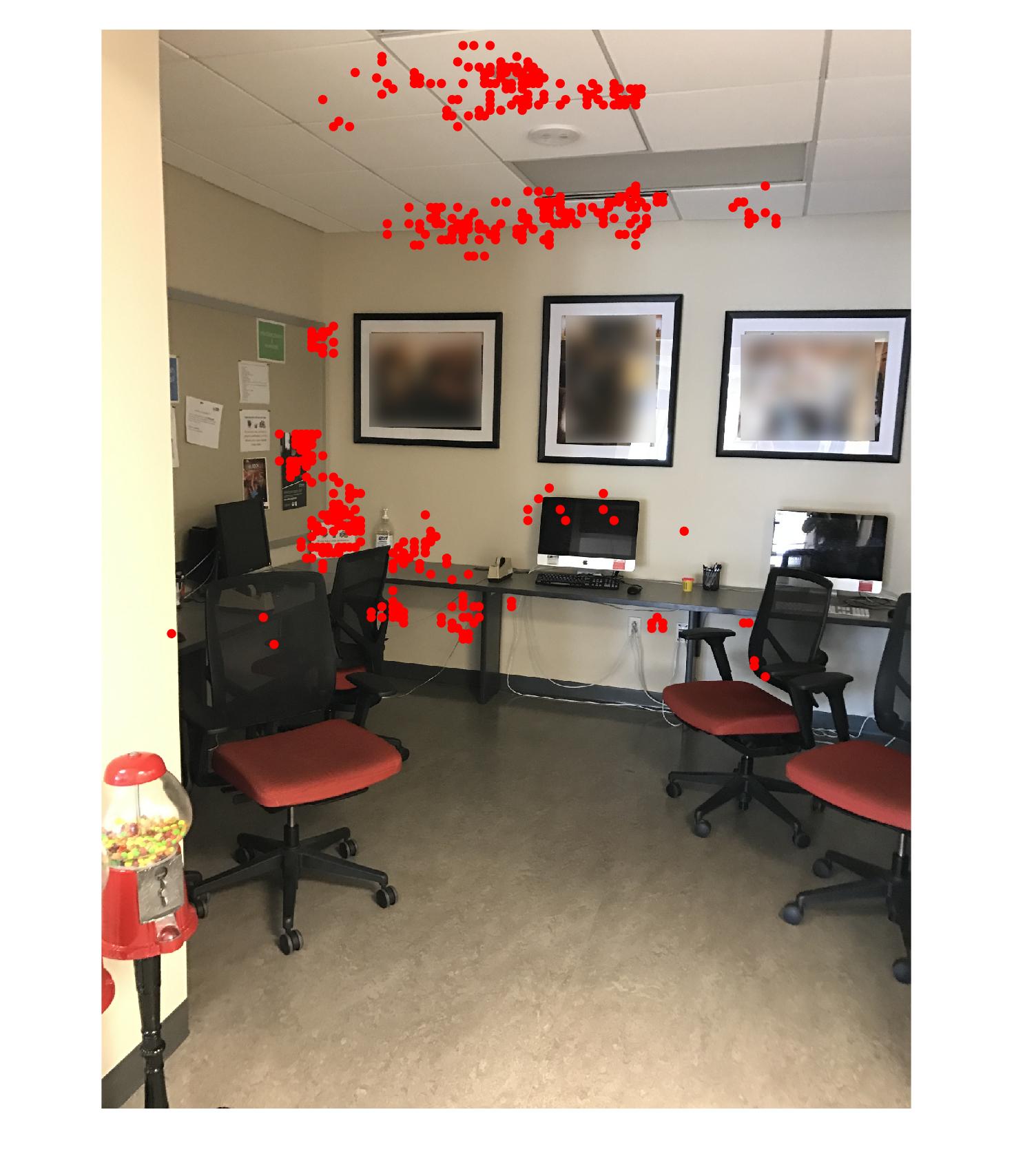}
        \includegraphics[height=3cm]{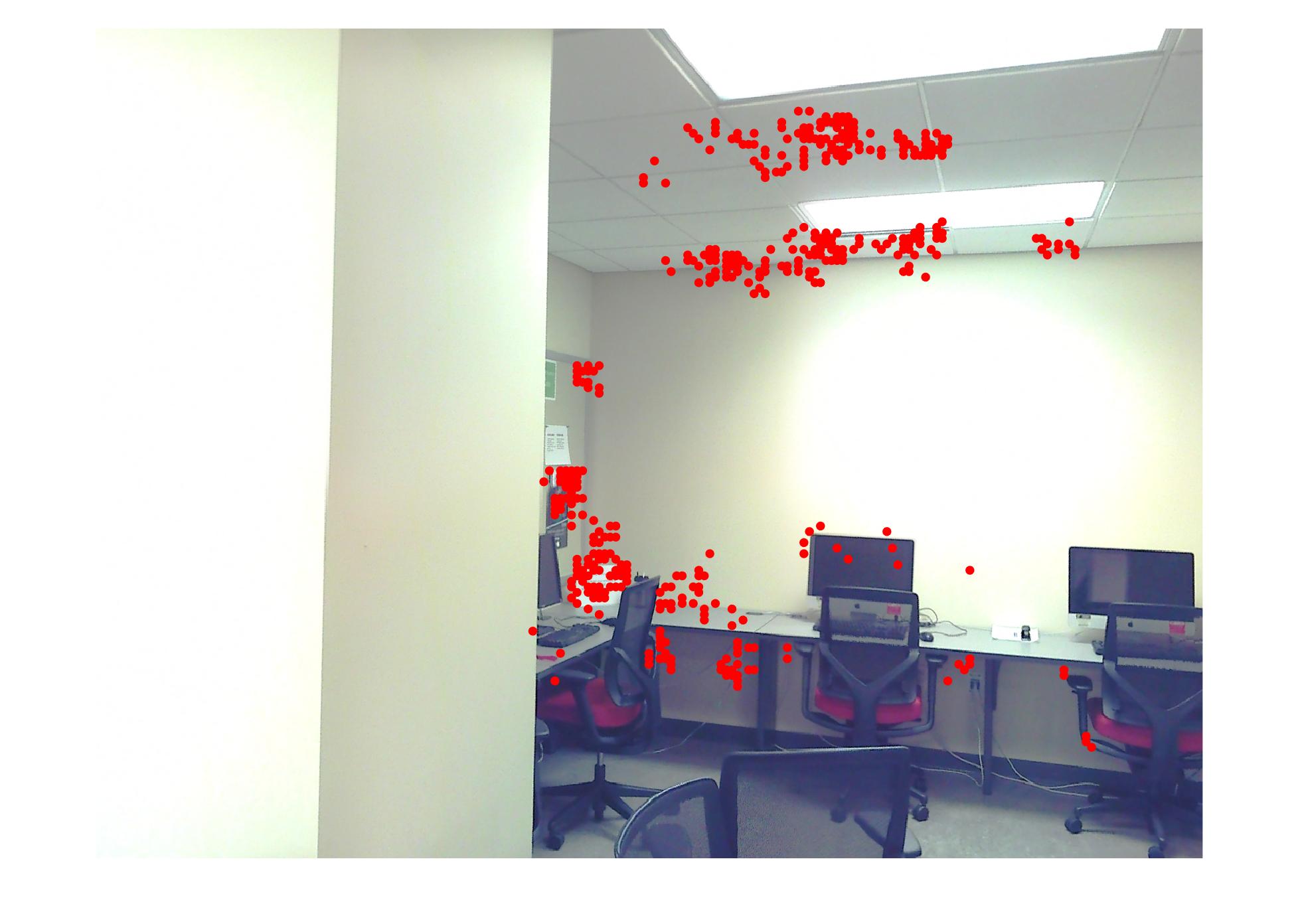}
    \end{subfigure}

    \begin{subfigure}[t]{0.48\textwidth}
        \centering
        \includegraphics[height=3cm]{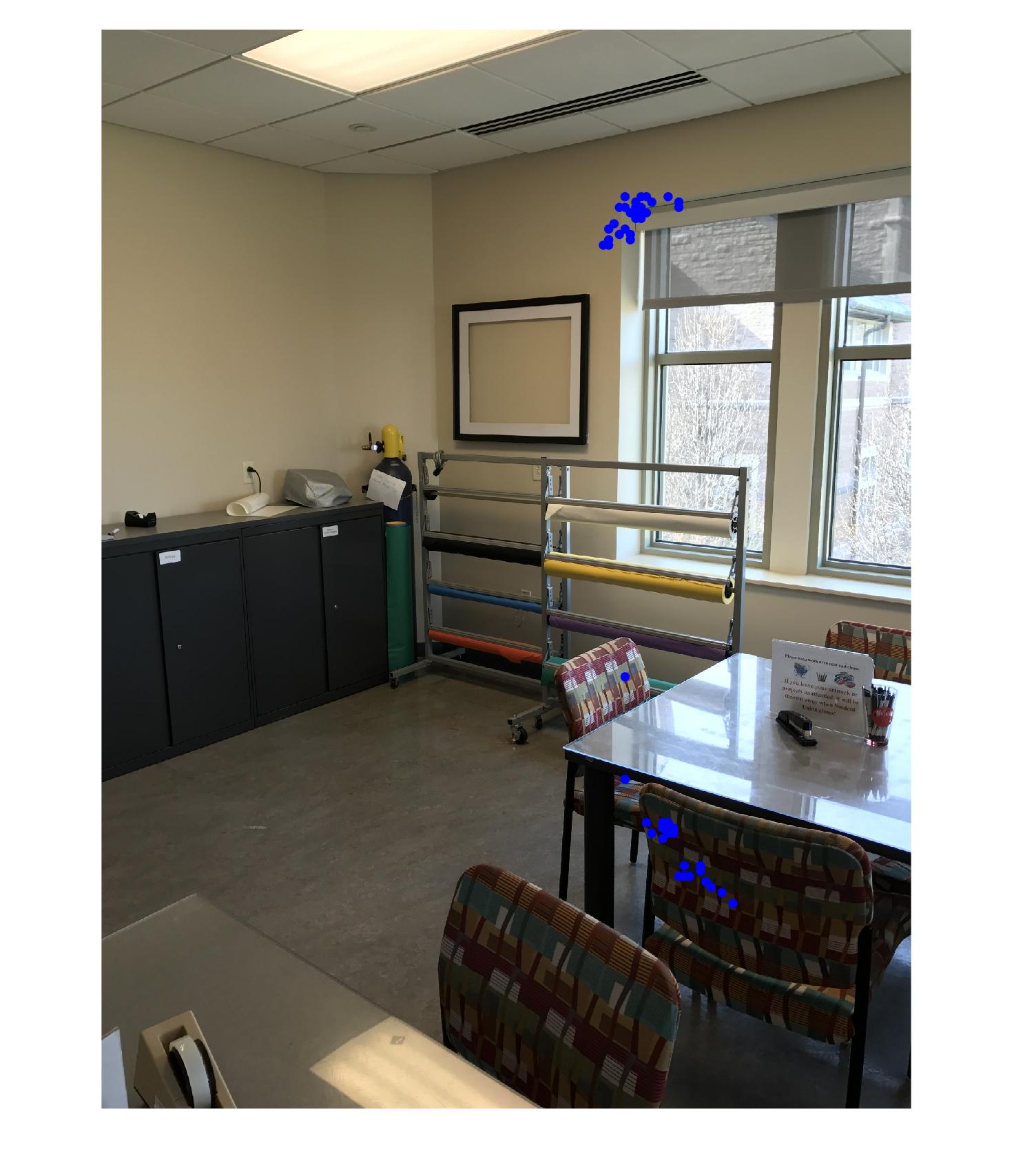}
        \includegraphics[height=3cm]{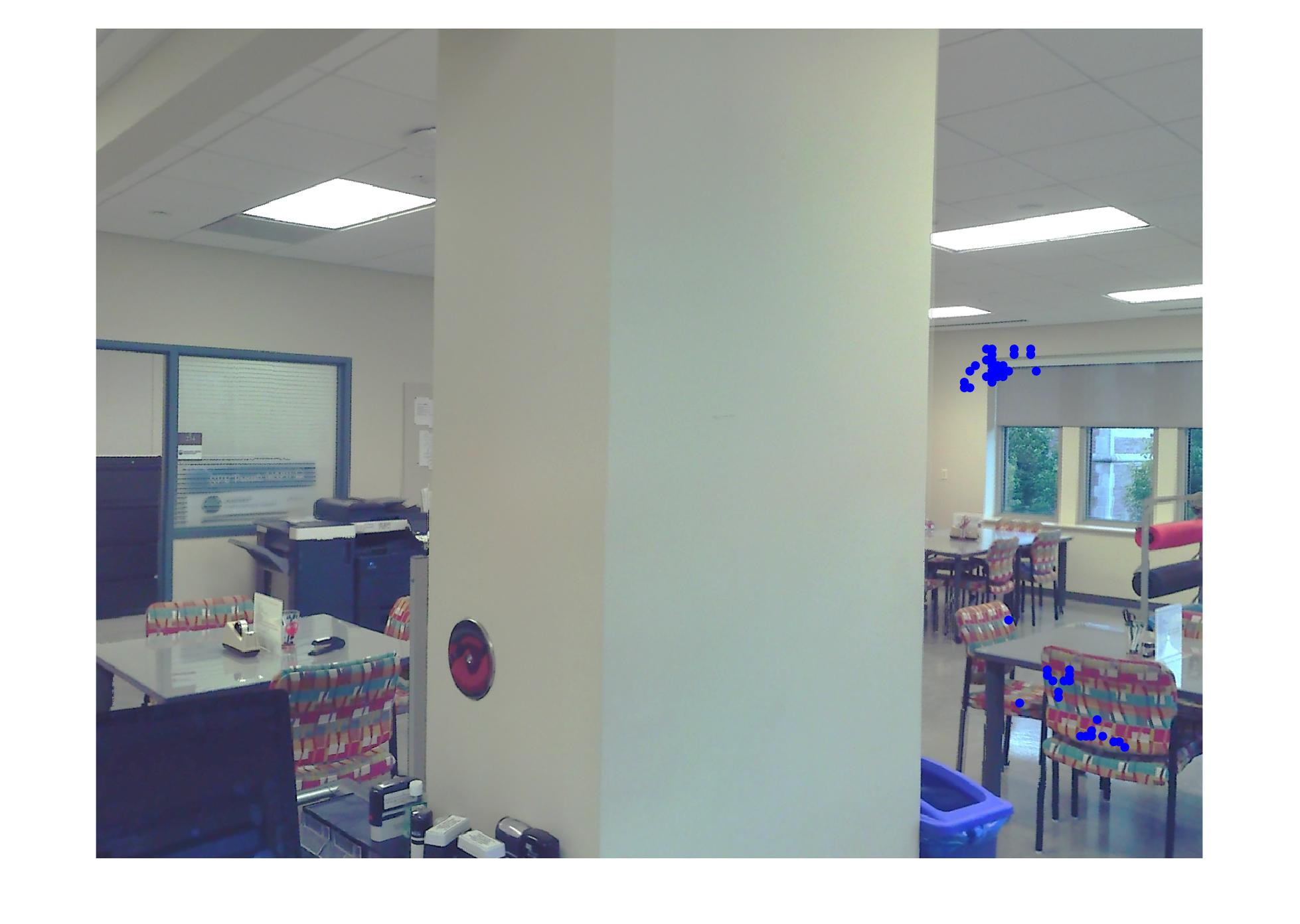}
    \end{subfigure}
    \begin{subfigure}[t]{0.48\textwidth}
        \centering
        \includegraphics[height=3cm]{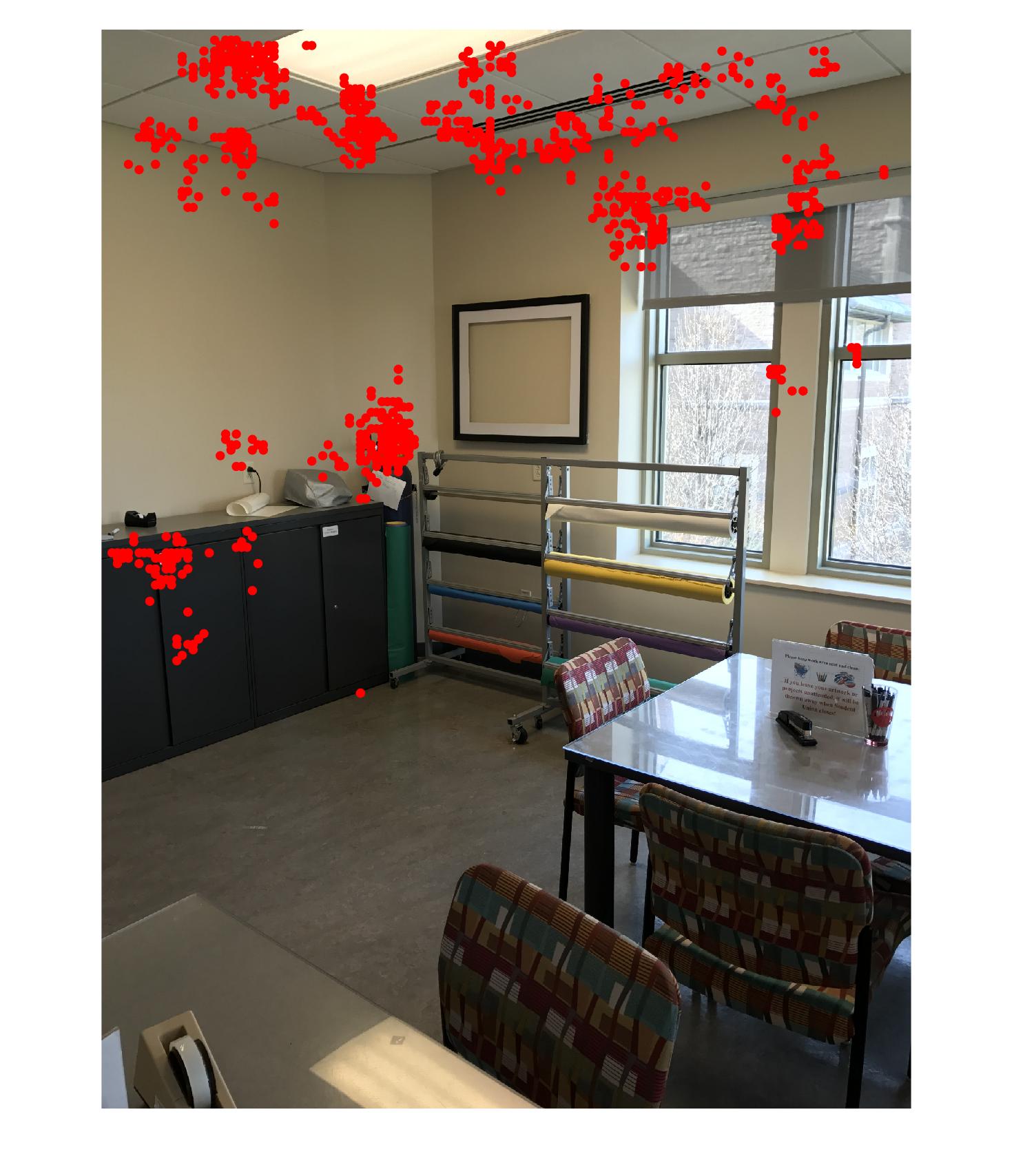}
        \includegraphics[height=3cm]{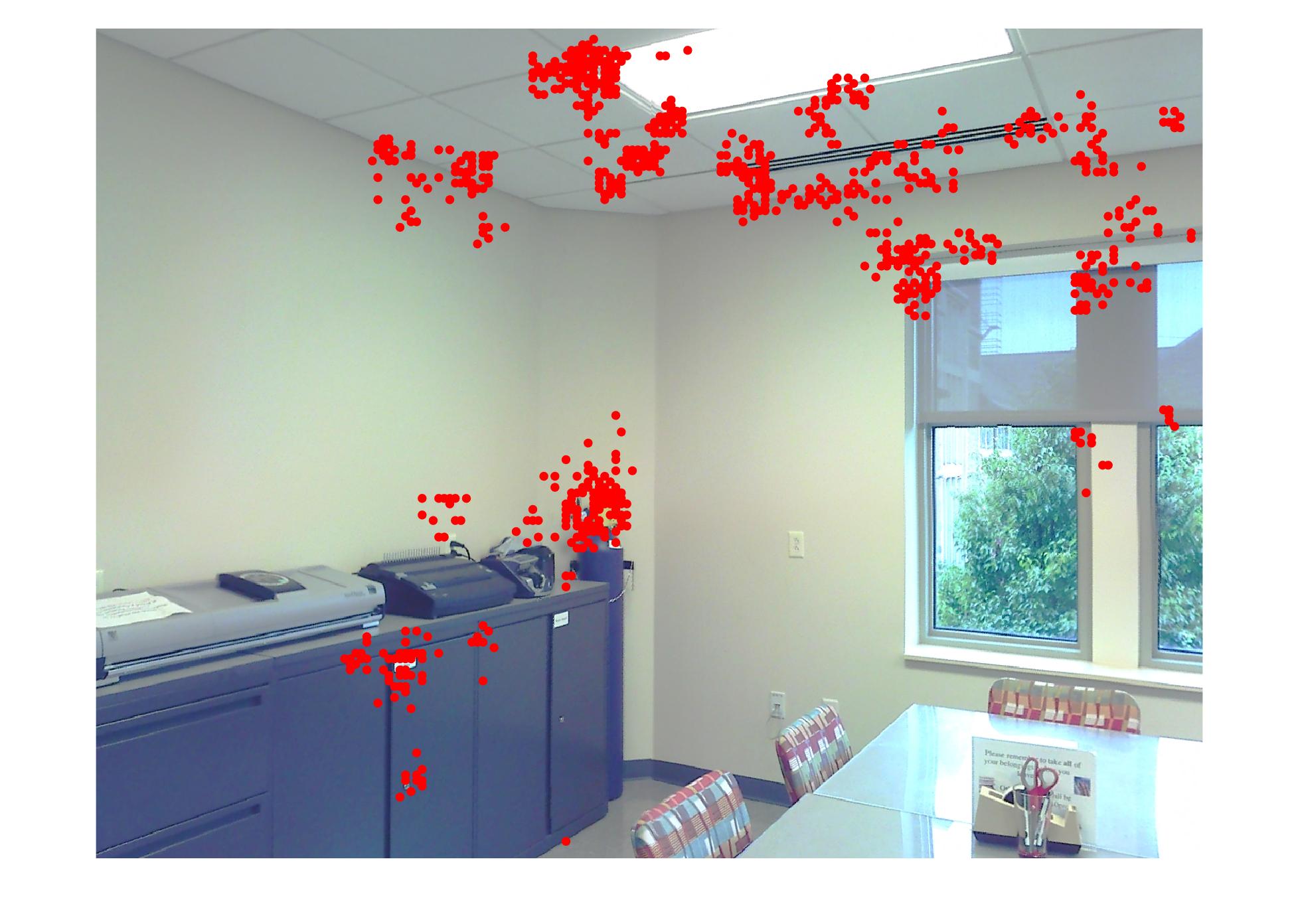}
    \end{subfigure}

    \begin{subfigure}[t]{0.48\textwidth}
        \centering
        \includegraphics[height=3cm]{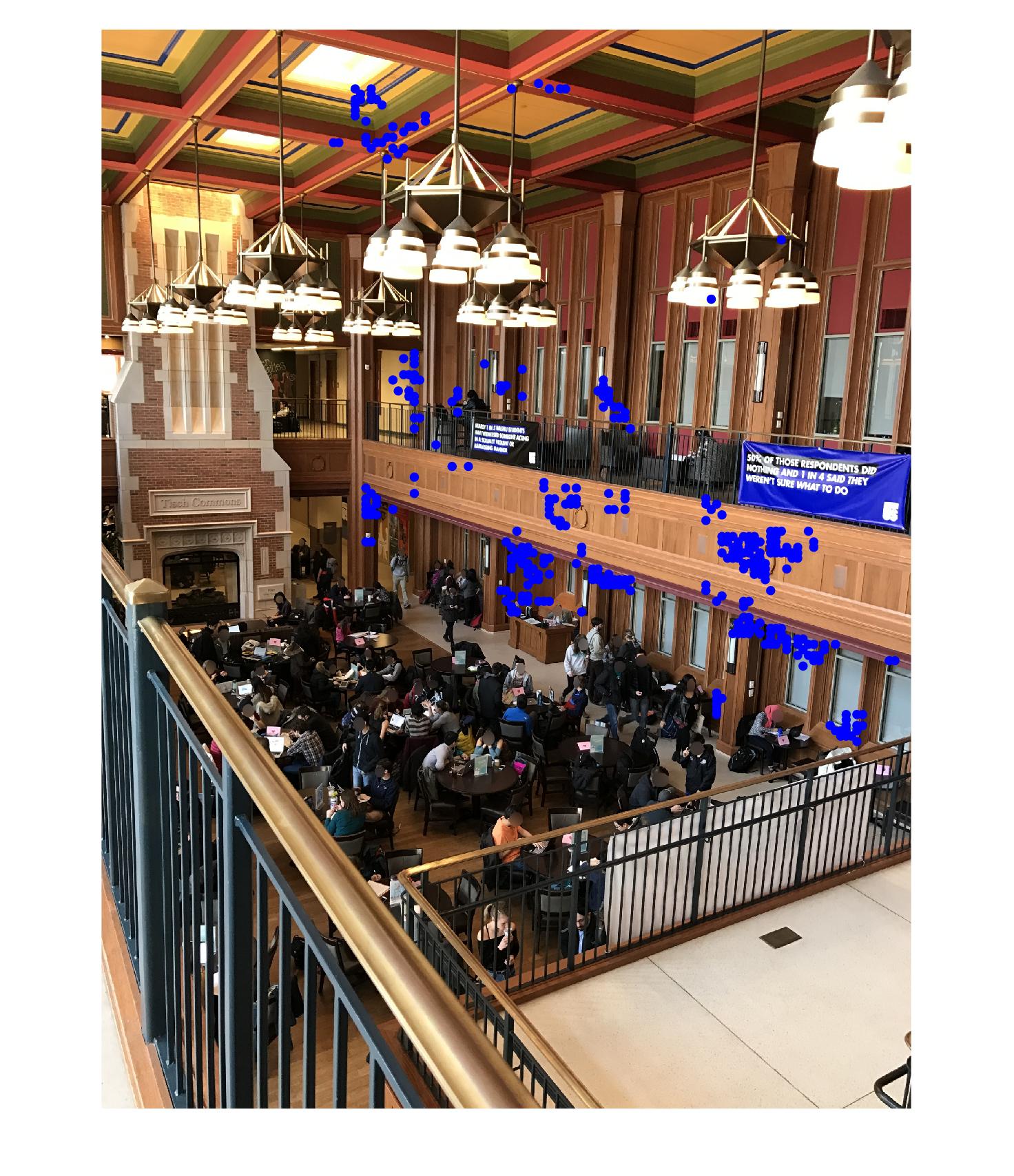}
        \includegraphics[height=3cm]{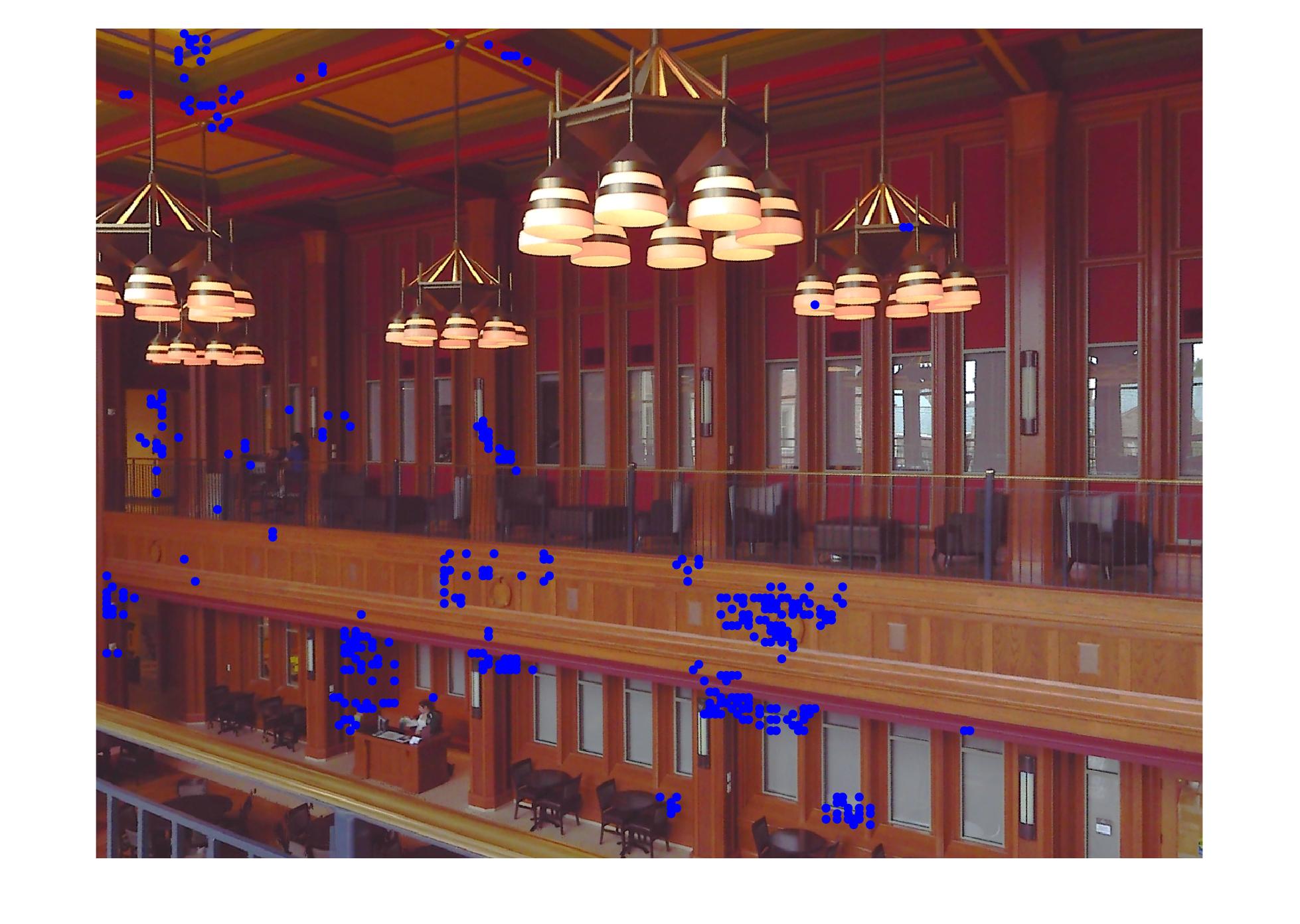}
    \end{subfigure}
    \begin{subfigure}[t]{0.48\textwidth}
        \centering
        \includegraphics[height=3cm]{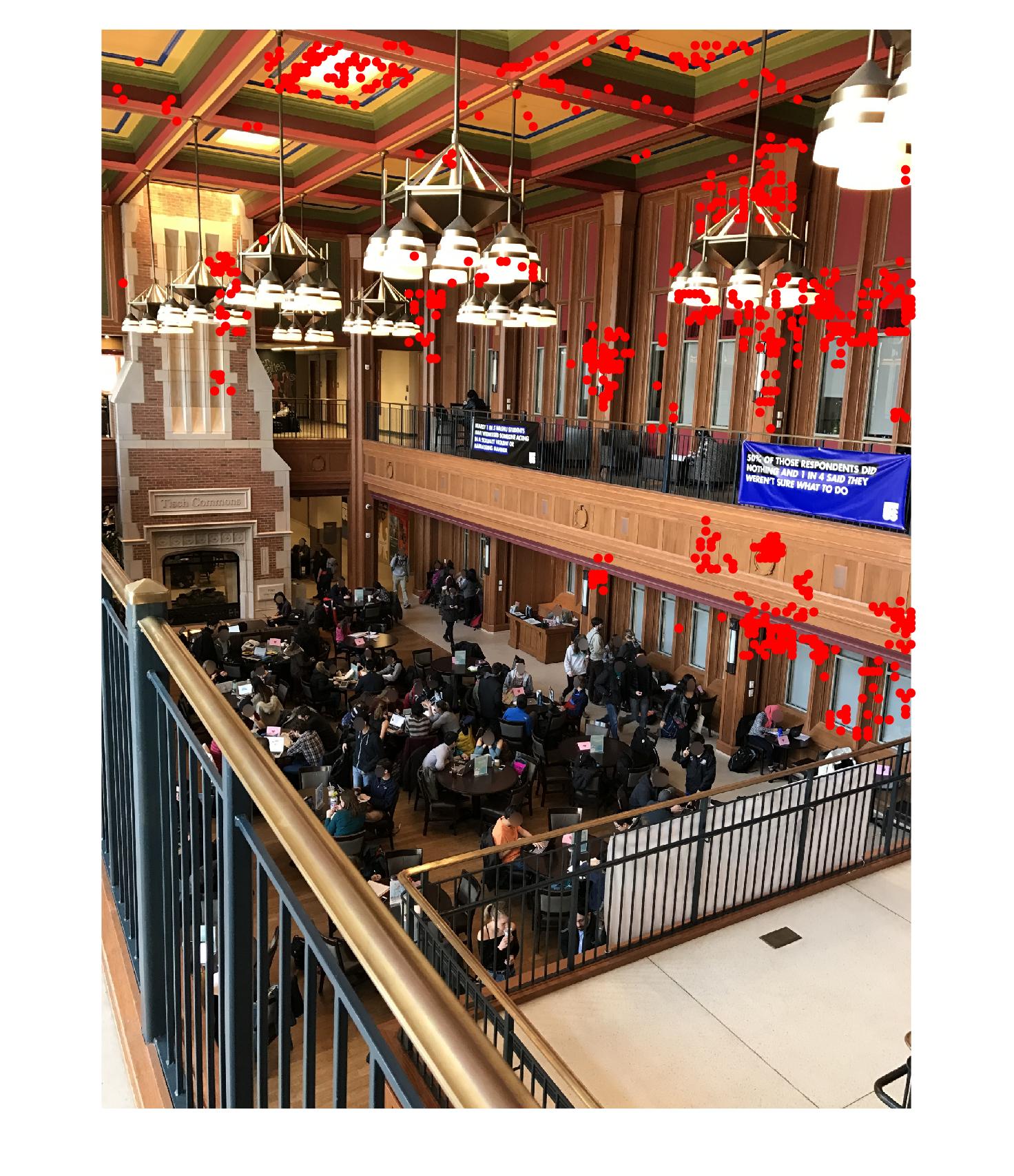}
        \includegraphics[height=3cm]{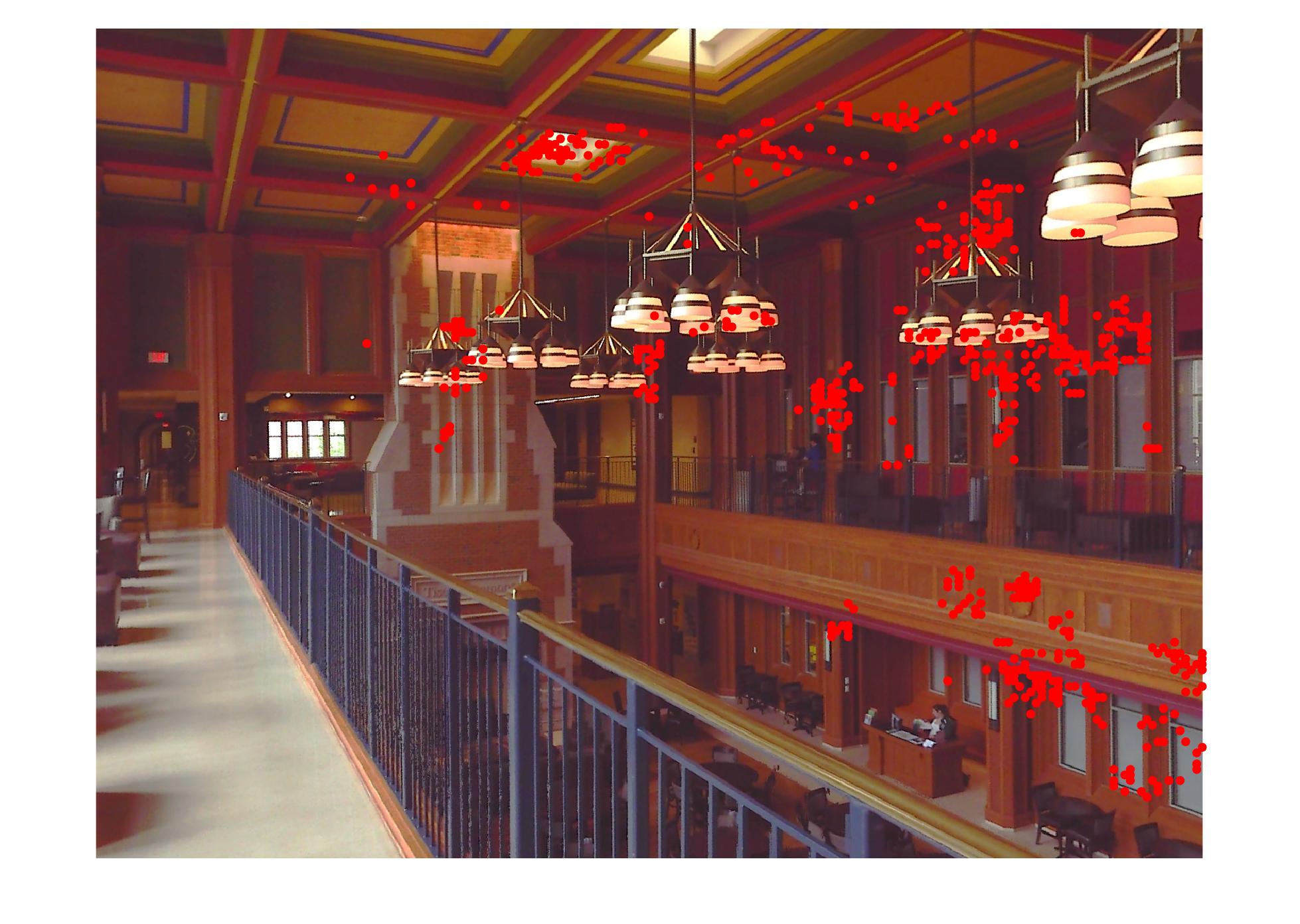}
    \end{subfigure}
    \caption{Qualitative comparison of best candidate poses with InLoc (left, inliers highlighted in blue) and our method (right, inliers highlighted in red), both including PV. \textbf{First row:} InLoc error: $\SI{32}{\meter}$, $\SI{3.9}{\degree}$. Our error: $\SI{0.09}{\meter}$, $\SI{3.7}{\degree}$. \textbf{Second row:} InLoc error: $\SI{5.3}{\meter}$, $\SI{14.3}{\degree}$. Our error: $\SI{0.24}{\meter}$, $\SI{2.9}{\degree}$. \textbf{Third row:} InLoc error: $\SI{0.56}{\meter}$, $\SI{2.1}{\degree}$. Our error: $\SI{4.4}{\meter}$, $\SI{2}{\degree}$.}
    \label{fig:comparison}
\end{figure*}

Figure \ref{fig:comparison} shows some examples where our method can correctly compute the pose but InLoc cannot, as well a case where InLoc succeeds but our method fails. As can be noticed, those cases where InLoc pose verification fails have a smaller amount of inliers, covering only a small area of the image. For example, for the query image in the first row, InLoc matches a chair in the query image to a chair in the database image, leading to a translation error of 32~m. Our method, where the pose verification score is weighted with the number and spatial distribution of inliers, penalizes this candidate, as the inliers only occupy a small area (coverage score 6\%) and chooses a better candidate, leading to a translation error of only 9~cm.

Despite that our method without PV is still worse than InLoc+PV, it is good to notice that the proposed approach does not require a 3D model. As it still gives a significant improvement to the basic pipeline without PV, it is an appealing alternative for those situations where dense 3D models are not available. \textit{Furthermore, our model without PV is also extremely fast to compute, being thus more appropriate for real-time applications than PV.} While PV takes on average $\SI{47}{\second}$ per query image, our verification score without PV can be computed in milliseconds. 

\subsection{Generalization to Other Datasets}
In the experiments so far the InLoc dataset was used for both training and testing. Similar to the number of inliers, a good confidence estimator should however be independent of the dataset, algorithm or error threshold used. For this reason, we investigated how our approach, trained on the InLoc dataset and algorithm with $\SI{1}{\meter}$ and $\SI{10}{\degree}$ error threshold, performs when applied to unforeseen datasets and pipelines. Particularly, we consider two outdoor visual localization datasets: Cambridge Landmarks and Aachen day/night and the coordinate regression based pipelines presented in \cite{li2019}.

\textbf{Cambridge Landmarks} contains several scenes from around Cambridge University \cite{kendall2015}. This dataset is not particularly challenging, as query and database images were taken under similar conditions, lacking e.g. strong differences in viewpoint or illumination. In these experiments, we consider the regression based pipeline presented in \cite{li2019}, where 3D coordinates of the pixels of the query image are computed and the so obtained 2D-3D correspondences are used to compute the camera pose with a P3P-RANSAC loop. As there is no image retrieval step, only the number of inliers and query image coverage score are available. Here, the pose is considered correct using an error threshold of 0.35~m and 5$^\circ$. As Table \ref{tab:ownCambridge} shows, our method still manages to discriminate successful and unsuccessful poses when applied to a completely different dataset, where the poses were obtained using a different pipeline and a different error threshold. Furthermore, our approach outperforms the number of inliers measure in most of the scenes. It should be noted that Cambridge Landmarks is a fairly simple dataset, where most competitive visual localization pipelines can reach accuracies over 90\%. As confidence estimation is particularly relevant for challenging datasets, the purpose of this experiment was \textit{not} to outperform the number of inliers measure, as both give good AUC values already. This experiment was meant to show that our confidence estimator, despite being trained with InLoc, is still valid with unforeseen datasets and pipelines.

\begin{table}[tb]
    \centering
    \caption{Performance of our InLoc trained algorithm with the  Cambridge Landmarks dataset.}
    \label{tab:ownCambridge}
    \begin{tabular}{c|c|c}
         Scene&AUC (inls count)&AUC (ours)  \\\hline\hline
         Great Court&80.08\%&83.59\%\\
         King's College&94.88\%&95.41\%\\
         Old Hospital&89.64\%&88.06\%\\
         Shop Facade&98.95\%&98.95\%\\
         St. Mary's Church&99.80\%&99.81\%\\\hline
         \textbf{Total}&\textbf{93.79\%}&\textbf{94.75\%}
    \end{tabular}
\end{table}

\textbf{Aachen} day/night dataset presents several images from the Aachen city during day and night time \cite{sattler2018}. The pipeline used by the authors of \cite{li2019} with this dataset is an approach, where both image retrieval and coordinates regression techniques are combined. Particularly, for a given query image, 10 candidate poses are produced. In their original work, the authors choose the best pose based on the number of inliers. Ground truths for the Aachen dataset are not available and hence the PR-curve cannot be computed. However, an online evaluation tool \cite{benchmarking} for computing the accuracy for the estimated poses, is available. The accuracies are computed for day scenes and night scenes separately. In our experiments, we use our confidence estimation algorithm to rerank the candidates of each query image. Since the considered pipeline has a retrieval step, we can use our full model with number of inliers and both coverage scores can be used. 
\begin{table}[tb]
    \centering
        \caption{Accuracies on Aachen using our method to choose the best candidate pose. The integration of our method into the visual localization pipeline of \cite{li2019} allows to reach better performances. Thresholds: $\SI{0.25}{\meter},\SI{2}{\degree}/\SI{0.5}{\meter},\SI{5}{\degree}/\SI{5}{\meter},\SI{10}{\degree}$ for day and $\SI{0.5}{\meter},\SI{2}{\degree}/\SI{1}{\meter},\SI{5}{\degree}/\SI{5}{\meter},\SI{10}{\degree}$ for night.}
    \label{tab:aachen}
    \begin{tabular}{c|c|c}
         Scene&Baseline accuracies [\%]&Our accuracies [\%]\\\hline\hline
         Day&$70.9 / 81.9 / 91.6$& $71.4 / 83.0 / 91.6$\\
         Night&$40.8 / 56.1 / 75.5$&$42.9 / 58.2 / 75.5$\\
    \end{tabular}
\end{table}

Table \ref{tab:aachen} shows the accuracies obtained with the baseline algorithm and our method. Ranking candidate poses with our method reaches higher accuracies at all thresholds for both day and night scenes, showing our method can be flexibly integrated into several visual localization pipelines to achieve a more accurate pose. Again, no dense 3D model nor heavy computations were needed to achieve the reported improvements.

\section{Conclusions}
\label{ch:conclusions}
In this paper we brought to attention a new research question, which has not been addressed sufficiently so far. Due to its ill-posed nature, pose estimation cannot always be successfully performed for an arbitrary query image. Hence, we developed a confidence estimation algorithm, that computes the probability of the estimated pose to be correct. In contrast to previous uncertainty modelling works, our approach allows to compare the confidence of different query images and is more robust than the number of RANSAC inliers. We have also shown that our metric is not dependent on the dataset nor on the error threshold used for training and can thus be flexibly integrated into different visual localization pipelines.

\section*{Ackowledgments}
This work was partially funded by the Academy of Finland project 309903 CoEfNet. We acknowledge the computational resources provided by the Aalto Science-IT project and CSC ${\text -}$ IT Center for Science, Finland.

{\small
\bibliographystyle{IEEEtranS}
\bibliography{egbib}
}

\end{document}